%% file: direct_radar_gyro.tex
\documentclass[conference]{IEEEtran}
\usepackage{times}

\usepackage[numbers]{natbib}
\usepackage{multicol}
\usepackage[bookmarks=true]{hyperref}
\usepackage{booktabs}
\usepackage[weather]{ifsym}

\input{macros}
\usepackage{soul}

\begin{document}

\title{DRO: Doppler-Aware Direct Radar Odometry}

\author{\authorblockN{Cedric Le Gentil\authorrefmark{1},
Leonardo Brizi\authorrefmark{2},
Daniil Lisus\authorrefmark{1}, 
Xinyuan Qiao\authorrefmark{1},
Giorgio Grisetti\authorrefmark{2} and 
Timothy D. Barfoot\authorrefmark{1}}
\authorblockA{\authorrefmark{1}Autonomous Space Robotics Lab\\
University of Toronto Institute for Aerospace Studies (UTIAS),
Toronto, Ontario, Canada\\ Email: \{cedric.legentil; daniil.lisus; samxinyuan.qiao; tim.barfoot\}@utoronto.ca}
\authorblockA{\authorrefmark{2}Department of Computer, Control, and Management
Engineering ``Antonio Ruberti"\\
Sapienza University of Rome\\
Email: \{brizi; grisetti\}@diag.uniroma1.it}
}

\maketitle

\begin{abstract}
A renaissance in radar-based sensing for mobile robotic applications is underway.
Compared to cameras or lidars, millimetre-wave radars have the ability to `see' through thin walls, vegetation, and adversarial weather conditions such as heavy rain, fog, snow, and dust.
In this paper, we propose a novel SE(2) odometry approach for spinning frequency-modulated continuous-wave radars.
Our method performs scan-to-local-map registration of the incoming radar data in a direct manner using all the radar intensity information without the need for feature or point cloud extraction.
The method performs locally continuous trajectory estimation and accounts for both motion and Doppler distortion of the radar scans.
If the radar possesses a specific frequency modulation pattern that makes radial Doppler velocities observable, an additional Doppler-based constraint is formulated to improve the velocity estimate and enable odometry in geometrically feature-deprived scenarios (e.g., featureless tunnels).
Our method has been validated on over $250 \, \si{\km}$ of on-road data sourced from public datasets (Boreas and MulRan) and collected using our automotive platform.
With the aid of a gyroscope, it outperforms state-of-the-art methods and achieves an average relative translation error of 0.26\% on the Boreas leaderboard.
When using data with the appropriate Doppler-enabling frequency modulation pattern, the translation error is reduced to 0.18\% in similar environments.
We also benchmarked our algorithm using 1.5 hours of data collected with a mobile robot in off-road environments with various levels of structure to demonstrate its versatility.
Our real-time implementation is publicly available:
\url{https://github.com/utiasASRL/dro}.
\end{abstract}

\IEEEpeerreviewmaketitle

\section{Introduction}

\begin{figure}
    \centering
    \begin{tikzpicture}
        \node [inner sep = 0em, outer sep = 0em] (boreas) {\includegraphics[width=0.49\columnwidth, clip,trim=0.0cm 0.0cm 0.0cm 0.0cm]{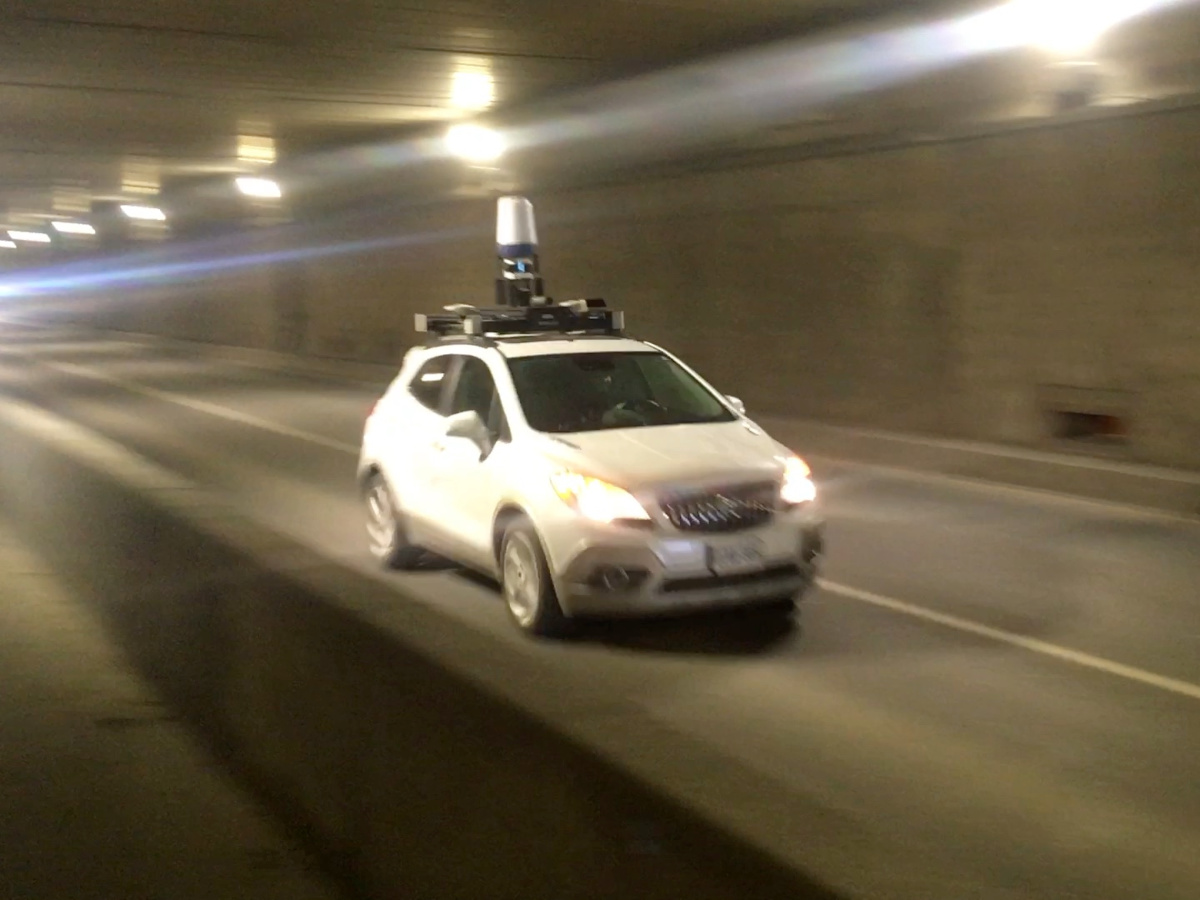}};
        \node [right=0.1cm of boreas, inner sep = 0em, outer sep = 0em] {\includegraphics[width=0.49\columnwidth, clip,trim=0.0cm 0.0cm 0.0cm 0.0cm]{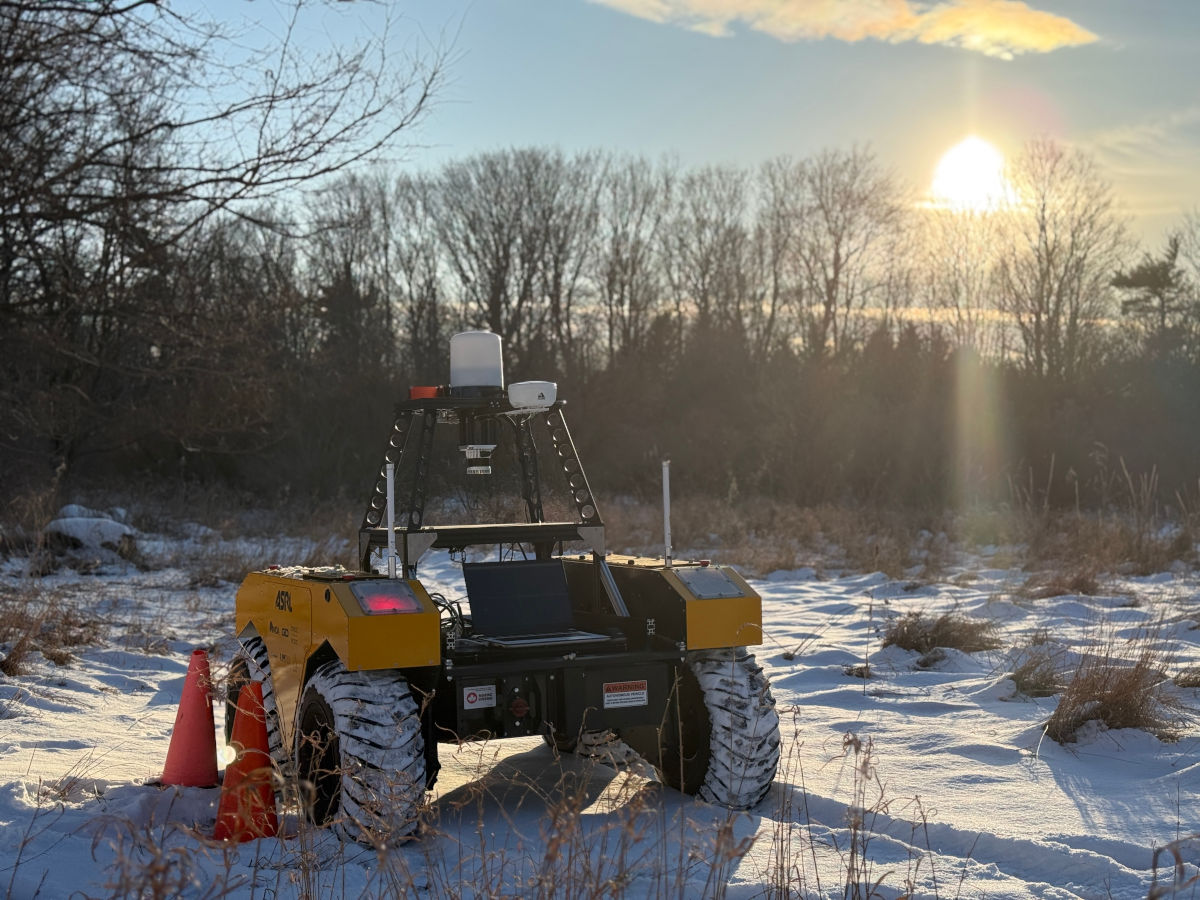}};
        \node [below=0.2cm of boreas, inner sep = 0em, outer sep = 0em] (boreastraj){\includegraphics[width=0.49\columnwidth, clip,trim=0.0cm 0.0cm 0.0cm 0.0cm]{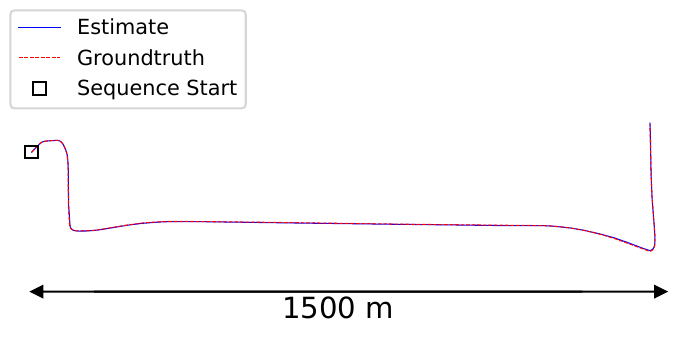}};
        \node [right=0.1cm of boreastraj, inner sep = 0em, outer sep = 0em](offroadtraj)  {\includegraphics[width=0.49\columnwidth, clip,trim=0.0cm 0.0cm 0.0cm 0.0cm]{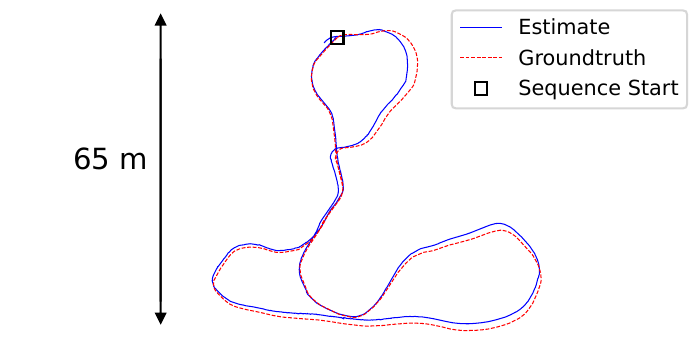}};
        \node [above=-1.1cm of boreastraj, inner sep = 0em, outer sep = 0em] {Tunnel};
        \node [right=-1.4cm of offroadtraj, inner sep = 0em, outer sep = 0em] {Offroad};
    \end{tikzpicture}
    \caption{This paper presents a Direct Radar Odometry (DRO) method that provides state-of-the-art performance in a wide range of environments, including unstructured ones (right) and geometrically degenerated ones such as featureless tunnels (left).}
    \label{fig:teaser}
\end{figure}

From air traffic control to weather forecasting via marine navigation, radar sensing is a cornerstone of modern societies.
During the early days of mobile robotics, radars were used to detect specific landmarks and perform \ac{slam}~\cite{dissanayake2001slam}.
While the advent of lidar technologies and the ubiquity of cameras drove the state estimation research community away from radar sensing~\cite{cadena2016past}, we are currently experiencing a `new wave' of interest toward this modality~\cite{harlow2024newwave}.
This interest is partly driven by the robustness of radars with respect to adversarial weather conditions.
In this paper, we present a novel algorithm for radar-based odometry and benchmark its performance across different datasets in both automotive and off-road robotic scenarios as shown in Fig.~\ref{fig:teaser}.

The working principle of radar sensing consists of emitting radio waves and analyzing the signal returned through various reflections of the waves from objects present in the environment to collect geometric information about the surroundings.
This work leverages data from \ac{mmwave} radars that use electromagnetic waves with a length scale that is in the order of magnitude of one millimetre.
Waves in such frequency bands can penetrate through plastic, textiles, glass, etc.
Accordingly, in the context of mobile robotics, radar allows the system to sense its environment in a wide range of conditions, even under intensely adverse weather conditions such as fog, rain, and snow.
Like any wave, the radar-emitted signals are subject to the Doppler effect when the relative velocity between the system and part of its environment is not zero.
Thus, when using specific frequency patterns and processing, radar data can allow for the extraction of Doppler-based radial velocity information.
These benefits represent a competitive advantage of radars compared to other modalities, such as vision or commonly used lidars.
The proposed odometry method accounts for the Doppler-based distortion of the radar data and can integrate a Doppler-specific velocity constraint in its optimization, allowing it to robustly perform odometry even in geometrically feature-deprived environments, such as featureless tunnels (see Fig.~\ref{fig:teaser} (left)).

Two types of radar sensors are typically used in robotics applications: `fixed' phased-array radars and spinning radars.
In this work, we focus on spinning radars, as they provide a $360\si{\degree}$ \ac{fov} that contains dense intensity information from up to several hundred meters away.
This makes spinning radar enticing for navigational tasks such as odometry, localization, and object tracking, where a view of the entire surroundings is required for robustness and safety.
The most common spinning-radar-odometry approaches use the dense intensity information to extract a set of points or features \cite{cen2019, adolfsson2023cfear, burnett2024continuous, under_the_radar, hero_paper, orora, qiao2024radar}.
These approaches work well, with point-based methods currently being the \ac{sota} \cite{adolfsson2023cfear, li2024cfearpp}, but they require computationally costly data association, critical parameter tuning, and they ultimately discard a large amount of the gathered data in favour of sparse representations.
Methods that have used the entire scan directly for odometry have done scan-to-scan correlation matching on the intensity image ~\cite{park2020pharao}, sometimes requiring pre-processing using a deep learning network \cite{masking_by_moving, weston2022fast-mbym}.
However, these methods do not explicitly account for the motion and Doppler distortion of the radar data, and discretize the relative transform search space to lower the computational cost.
Additionally, the need for a learned processing component limits the transferability of these approaches to different environments.

In this work, we present the first Direct Radar Odometry (DRO) framework that accounts for the continuous motion and Doppler distortion of the incoming data in a principled way.
It can be aided by a yaw-axis gyroscope and does not rely on deep-learning techniques or hand-tuned discretization parameters.
As demonstrated on more than $250\,\si{\km}$ of data, it is applicable to a wide range of environments.
With the appropriate radar emission pattern, it can also perform in geometrically challenging scenarios that generally lead to degenerated state estimates.
The main contributions are as follows:
\begin{enumerate}
    \item The formulation of the first direct radar registration method that accounts for continuous motion and Doppler-based distortion in a gradient-based optimization that does not discretize the search space.
    \item A novel way to leverage the Doppler effect for velocity estimation from spinning radar without assuming that consecutive radar beams correspond to the same objects.
    \item Our method is capable of \ac{sota} radar-odometry performance when used with a gyroscope.
    \item A real-time GPU-based implementation. \footnote{\url{https://github.com/utiasASRL/dro}}
\end{enumerate}

\section{Related work}
A radar odometry system \cite{harlow2024newwave,nader2024survey,venon2022millimeter,cen2018precise}, like other odometry systems, aims to estimate the ego-motion of the sensor, in this case, a radar, thus determining the trajectory of the platform on which the sensor is mounted. The motion estimation process typically involves formulating an optimization problem, where the objective is to find the sensor's state and an environment model that maximizes the probability of observing the measured data. This principle is known as maximum-likelihood estimation.

In the last decade, radar odometry has been dominated by indirect approaches \cite{nader2024survey}, which rely on selecting salient points or features from radar measurements. Early works, such as \cite{callmer2011radarslamusingvisualfeatures}, started by leveraging computer vision approaches, such as SIFT \cite{lowe2004sift} features, from radar Cartesian images, while more recent state-of-the-art methods, such as CFEAR \cite{adolfsson2023cfear}, select points based on each individual azimuths of a radar signal.
There exist two main categories for feature extraction methods \cite{pprestonkrebs2024thefinerpoints}: signal-based and spatial-based extractors.
Signal-based methods identify points by applying thresholds to individual azimuths in the radar signal. A well-known example is the Constant False Alarm Rate (CFAR) method \cite{finn1968adaptivedm}, which processes each azimuth of a radar scan using a sliding window. The power readings within the window are used to calculate a detection threshold. Many variants of CFAR exist \cite{fernández2017detectorescfar, rohling1983radarcfar, gandhi1988analysisofcfar, zhao2001novel, zhang2013improvedswitching, Smith1997VICFARAN, blake1988OSCFAR}. Another signal-based extractor, K-strongest \cite{adolfsson2021CFEAR}, selects the K strongest power returns above a static threshold along each azimuth. In contrast, spatial-based extractors rely on extracting features and descriptors from radar scans transformed into Cartesian images, some of those leverage computer vision algorithms, such as ORB \cite{rublee2011ORB} or SIFT \cite{lowe2004sift}.

Indirect odometry systems, extracting points from the radar data, rely on matching these features. The data association step is one of the most critical building blocks of these systems.
Burnett et al. \cite{2021_Burnett} introduced motion-compensated RANSAC to ensure robust association.
Subsequently, they developed Hero \cite{hero_paper}, a method that leverages learned key-point locations, uncertainties, and descriptors to address outliers and facilitate matching within the ESGVI \cite{barfoot2020esgvi} batch state estimation framework.
In CFEAR \cite{adolfsson2023cfear}, the authors extract the surface points and demonstrate that the point-to-line and point-to-distribution perform better than point-to-point, as these metrics mitigate the problem of erroneous associations between individual points.
After an initial scan filtering step, Kung et al. \cite{Kung2021AND} extract sets of local Gaussian distributions and perform point-to-distribution registration.
Finally, some other systems exploit data-driven approaches to downscale noisy point contributions in the optimization problem \cite{Lisus2023PointingTW} or to select others suitable for the data association \cite{Aldera2019}.

Alternatively, direct methods do not need explicit data association.
While this kind of method has been well studied for lidar \cite{Corte2017AGF,DiGiammarino2022,DiGiammarino2023} and monocular odometry \cite{Engel2014LSDSLAMLD,Forster2014SVOFS,Mona2019,Delaunoy2014,Engel2016}, the literature on direct radar-based odometry is still somewhat limited and methods are slower than feature-based methods.
Barnes et al. \cite{masking_by_moving} proposed a brute-force approach that samples and selects rotations to maximize the dense correlation between radar scans. Similarly, Checchin et al. \cite{Checchin2009} and Park et al. \cite{park2020pharao} addressed the need for sampling by decoupling rotation and translation. They utilized the Fourier-Mellin Transform, maximizing phase correlation between log-polar and Cartesian images to estimate rotation and translation, respectively. A coarse-to-fine phase correlation refinement on Cartesian images was then applied to improve the translation estimate. While Barnes et al. \cite{masking_by_moving} achieved remarkable performance in odometry estimation, their method suffered from high computational complexity. To address this, Weston et al. \cite{weston2022fast-mbym} replaced the brute-force rotation search with an FFT-based approach, significantly reducing computation time at the cost of slightly lower accuracy. Overall, existing direct methods either decouple the estimation of rotation and translation or sub-sample the problem within a discrete search space. These strategies can result in susceptibility to local minima and poor scalability across different resolutions. In contrast, our approach optimizes the locally continuous SE(2) radar trajectory without discretizing the search space.

For both direct and indirect methods, motion distortion can significantly degrade system performance. In the literature, this issue has been addressed using continuous-time trajectory estimation \cite{anderson2013ransac}, \cite{Anderson2015,Anderson2015FullSTEAM, Furgale2015} for lidar \cite{Bosse2008, omar2023} and rolling-shutter cameras \cite{Hedborg2012RollingSB}. This work applies similar concepts to radar data while also correcting Doppler distortion in a continuous-time optimization.
The Doppler information channel provides valuable additional data that can be leveraged for odometry and ego-motion estimation. Several works have explored its integration with different models and estimation techniques to enhance motion estimation accuracy. Kellner et al. \cite{Kellner2013} demonstrated that automotive radar, combined with an Ackermann vehicle model, can estimate ego-motion by fitting least-squares to stationary point Doppler velocities identified via RANSAC.
In~\cite{Ghabcheloo2018}, a gyroscope is used to alleviate the need for an Ackermann platform.
Galeote-Luque et al. \cite{GaleoteLuque2024} extended this by using Doppler measurements for 3D odometry with RANSAC and a kinematic model. Kubelka et al. \cite{Kubelka2024} showed that integrating high-quality Doppler data with an IMU outperforms ICP in feature-deprived environments, and Lisus et al. \cite{lisus2025doppler} implemented a continuous-time approach with a white-noise-on-acceleration prior.
In~\cite{Rennie_Williams_Newman_DeMartini_2023}, the authors extend the deep-learning-based preprocessing from~\cite{masking_by_moving} and trained an additional network for per-scan velocity regression.
In contrast, our model-based approach does not require any training data and does not suffer from any generalization issues.

\section{Background}

\subsection{Gaussian process regression}

\ac{gp} regression \cite{rasmussen2006gp} is a probabilistic, non-parametric interpolation method.
Let us consider a signal $h(\abscissavec)$, with $\abscissavec \in \reals^D$, modeled with a \ac{gp} $h(\abscissavec) \sim \gp\left(0, \kernel{}{\abscissavec}{\abscissavec'}\right)$. The function $\kernel{}{\abscissavec}{\abscissavec'}$ is the so-called covariance kernel that characterizes the covariance between two instances of $h$ (at $\abscissavec$ and $\abscissavec'$).
Given noisy samples $\ordinate_i = h(\abscissavec) + \noise$, with $\noise \sim \gaussian\left(0,\sigma\right)$ and $i=(1,\cdots, N)$, \ac{gp} regression aims at inferring the signal $h$ and its variance at any new input location $\abscissavec^*$.
The definition of a \ac{gp} corresponds to the following multivariate normal distribution:
\begin{equation}
    \label{eq:multivariate}
    \begin{bmatrix}
    \ordinatevec
    \\
    h(\abscissavec^*)
    \end{bmatrix}
    \sim
    \gaussian
    \left(
    \begin{bmatrix}
        \mathbf{0}\\ 0
    \end{bmatrix}
    ,
    \begin{bmatrix}
        \kernelmat{}{\abscissamat}{\abscissamat} + \sigma^2\identitymat & \kernelvec{}{\abscissamat}{\abscissavec^*}
        \\
        \kernelvec{}{\abscissavec^*}{\abscissamat} & \kernel{}{\abscissavec^*}{\abscissavec^*}
    \end{bmatrix}
    \right),
\end{equation}
where $\kernelvec{}{\abscissavec^*}{\abscissamat} = \left(\kernelvec{}{\abscissamat}{\abscissavec^*}\right)^\top = \begin{bmatrix}\kernel{}{\abscissavec^*}{\abscissavec_1}&\cdots&\kernel{}{\abscissavec^*}{\abscissavec_N}\end{bmatrix}$, $\kernelmat{}{\abscissamat}{\abscissamat} = \begin{bmatrix} \kernelvec{}{\abscissamat}{\abscissavec_1} & \cdots&\kernelvec{}{\abscissamat}{\abscissavec_N}\end{bmatrix}$, and $\ordinatevec = \begin{bmatrix} \ordinate_1&\cdots&\ordinate_N\end{bmatrix}$.
By conditioning \eqref{eq:multivariate} with respect to the noisy observations we obtain the inference of $h$ at $\abscissavec^*$ as
\begin{equation}
    \label{eq:gp_infer}
    \begin{aligned}
        \hat{h}(\abscissavec^*) &= \kernelvec{}{\abscissavec^*}{\abscissamat} \left(\kernelmat{}{\abscissamat}{\abscissamat} + \sigma^2\identitymat\right)^{-1} \ordinatevec,
        \\
        \text{var}\left(\hat{h}(\abscissavec^*)\right) &= \kernel{}{\abscissavec^*}{\abscissavec^*} - \kernelvec{}{\abscissavec^*}{\abscissamat} \left(\kernelmat{}{\abscissamat}{\abscissamat} + \sigma^2\identitymat\right)^{-1}\kernelvec{}{\abscissamat}{\abscissavec^*}.
    \end{aligned}
\end{equation}

\subsection{Radar data and Doppler effect}
\label{sec:doppler_radar}

Let us consider the type of radars commonly used in the automotive and robotic industries: \ac{fmcw} radars.
As their name suggests, these radars are continuously emitting frequency-modulated electromagnetic waves.
Thus, instead of directly measuring the time between the emission and reception of waves, the distance to an electromagnetic reflector corresponds to a frequency difference $\Delta f$ between the signal transmitted and received.
For the sake of simplicity, the following explanation will consider a single reflector at a certain distance $r$ from the radar.
In reality, the \ac{fmcw} radars provide a reflection score/value for many `range bins' along the emission beam.

\begin{figure}
    \centering
    \begin{tikzpicture}
        \node [inner sep = 0em, outer sep = 0em] {\includegraphics[width=0.99\columnwidth, clip,trim=0.2cm 0.2cm 0.2cm 0.2cm]{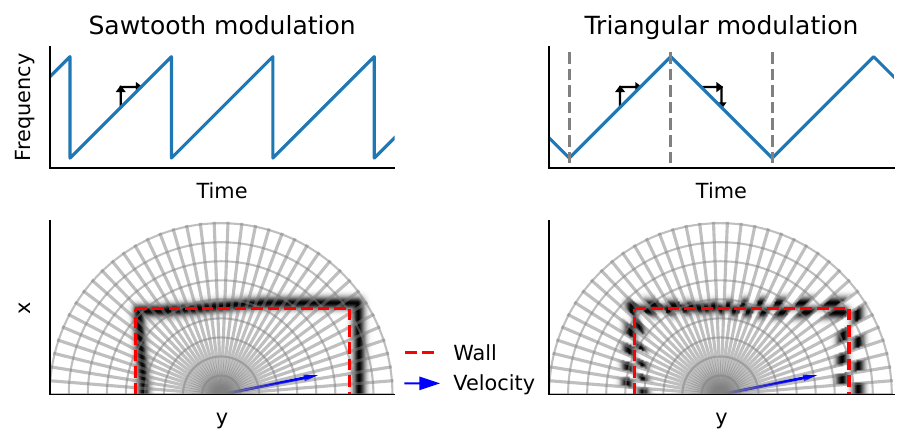}};
        \node at (-3.35,1.65) {\scriptsize Slope $s$};
        \node at (1.62,1.65) {\scriptsize Slope $s$};
        \node at (2.65,1.65) {\scriptsize Slope -$s$};
        \node at (1.65,0.65) {\tiny \color{gray} up-chirp};
        \node at (2.63,0.65) {\tiny \color{gray} down-chirp};
    \end{tikzpicture}
    \caption{Impact of different frequency modulation patterns (top) and Doppler effect on the data collected with spinning \ac{fmcw} radars (bottom).}
    \label{fig:radar_doppler}
\end{figure}
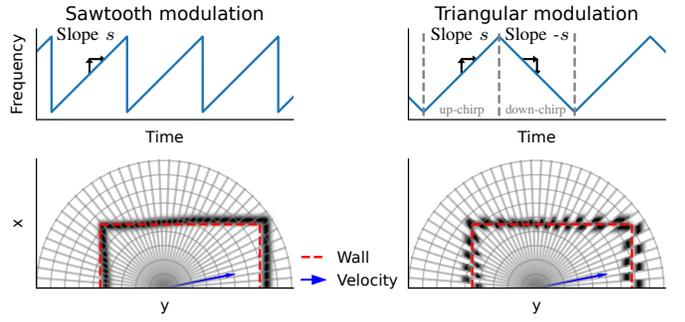

Considering linear modulation of the frequency with a slope $s$ as shown in Fig.~\ref{fig:radar_doppler}, the distance $r$ between the radar and reflector is
\begin{equation}
    \label{eq:range_meas}
    \tilde{r} = \frac{c \Delta f}{2s},
\end{equation}
where $c$ is the speed of light, and the divider $2$ accounts for the fact that $\Delta f$ represents twice the distance $r$ due to the time for the wave to reach the reflector and return.
This relationship is only true ($\tilde{r} = r$) if the relative velocity between the sensor and the reflector is null.
In the presence of relative motion, the measured $\Delta f$ is impacted by the Doppler effect that compresses or expands the electromagnetic wave, thus resulting in $\Delta f = \Delta f_t + \Delta f_d$, with $\Delta f_t$ being the shift corresponding to the waves travelling time, and $\Delta f_d$ the Doppler-induced frequency shift.
This shift $\Delta f_d$ is proportional to the signal frequency $\frac{1}{\lambda_t}$ and the velocity along the emission beam, denoted `radial velocity' $u$, as 
\begin{equation}
    \Delta f_d = \frac{2u}{\lambda_t}.
\end{equation}
Accordingly, the `range' measurement provided by \eqref{eq:range_meas} is perturbed by the Doppler effect
\begin{equation}
    \tilde{r} = \frac{c\Delta f_t}{2s} + \frac{c\Delta f_d}{2s} = r + r_d(u).
\end{equation}

When considering a frequency modulation with a \emph{sawtooth} pattern and multiple measurements of the same reflector, we cannot dissociate $\Delta f_t$ and $\Delta f_d$ from $\Delta f$.
However, if we use a triangle pattern consisting of a succession of `up-chirps' and `down-chirps' as illustrated in Fig.~\ref{fig:radar_doppler}, two measurements of the same reflector allow for the recovery of the radial velocity:
\begin{equation}
    \tilde{r}_{\up} - \tilde{r}_{\down} = \frac{c\Delta f_t}{2s} + \frac{c\Delta f_d}{2s} - \frac{-c\Delta f_t}{-2s} - \frac{c\Delta f_d}{-2s}
\end{equation}
where the components in $\Delta f_t$ cancel out resulting in
\begin{equation}
    \label{eq:delta_range}
    \Delta\hat{r} = \tilde{r}_{\up} - \tilde{r}_{\down} = \frac{2uc}{s\lambda_y} = \beta u.
\end{equation}

In reality, \ac{fmcw} radars do not only expect one return per beam but collect `reflectivity' information through a wide range of discrete ranges through a fast-Fourier transform of the received signal.
Thus, the data from such radars can be interpreted as intensity/reflectivity images, in which rows correspond to different beam azimuths and columns to different ranges.
Fig.~\ref{fig:radar_doppler} illustrates the effect of the Doppler effect on \ac{fmcw} data.
The data collected with a moving radar using a triangular pattern presents an alternating shift between each row of the scans, while all the rows are shifted according to the sign of the radial velocity for sawtooth radars.
Depending on the radar model, the modulation pattern can be changed or not in the firmware.
Thus, not all spinning \ac{fmcw} radars enable the extraction of Doppler-based velocities.

\section{Method}

The proposed method addresses the issue of radar odometry optionally aided by a gyroscope.
The main concept is the direct use of \ac{fmcw} intensity data without any point cloud or feature extraction, or data association.

\subsection{Problem statement and overview}

Let us consider a 2D \ac{mmwave} \ac{fmcw} spinning radar and a rigidly mounted gyroscope.
Without loss of generality, the extrinsic calibration (rotation matrix) between the two sensors is assumed to be known and will not appear in the following derivations.
The gyroscope provides measurements $\gyr{i}$, at times $\imutime{i}$, of the system's angular velocity around the spinning axis of the radar.
The radar data is assumed to be collected in scans covering the 360${}^\circ$ field-of-view.
Each scan contains information collected from $N$ beams, each associated with an azimuth $\az{n}$ and timestamp $\aztime{n}$, and each beam consists of the electromagnetic reflection intensity $\intensity{nm}$ at $M$ different ranges $\range{m}$.
Accordingly, a radar scan can be interpreted as a polar intensity image or a set of tuples $\{\az{n}, \range{m}, \intensity{nm}\}$.

\begin{figure*}
    \centering
    \begin{tikzpicture}
        \node[inner sep=0,outer sep=0]{\includegraphics[width=0.99\linewidth, clip]{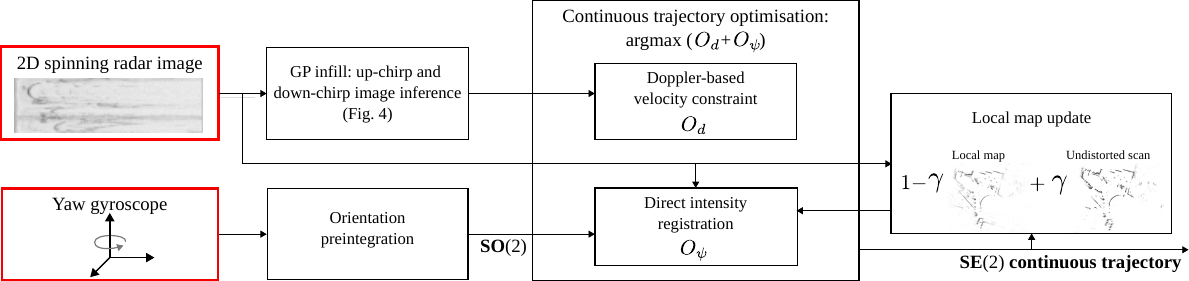}};
    \end{tikzpicture}
    \caption{Overview diagram of the proposed direct radar odometry method. To leverage the Doppler-based velocity constraint, the radar must use a triangular frequency modulation pattern. Note that $\objective{\intensity{}}$ accounts for motion and Doppler-induced distortion.}
    \label{fig:overview}
\end{figure*}

The rotation, position, and velocity of the radar reference frame with respect to an Earth-fixed frame are described with continuous functions $t \in \reals \mapsto \rot{\world}{t} \in SO(2)$, $\pos{\world}{t} \in \reals^2$, and $\vel{\world}{t} \in \reals^2$, respectively.
The details of the various motion models used to obtain $\rot{\world}{t}$, $\pos{\world}{t}$, and $\vel{\world}{t}$ from a finite set of state variables $\state$ will be discussed in Section~\ref{sec:motion_models}.
The proposed method aims to estimate the state variables $\state$, thus the sensor's trajectory, during the collection time of the last radar scan using the radar intensity data and optionally the angular velocity from the gyroscope.
The estimation is done with the numerical optimization of a combination of objective functions as
\begin{equation}
    \label{eq:optimisation}
    \state^* = \underset{\state}{\text{argmax}}\quad  \objective{\intensity{}} + \objective{d},
\end{equation}
with $\objective{\intensity{}}$ an objective function that relates to the direct registration of the radar intensity data (cf., Section~\ref{sec:direct_objective}, and $\objective{d}$ to the Doppler effect (cf., Section~\ref{sec:doppler_objective}).
Fig.~\ref{fig:overview} provides an overview of the proposed method.
As detailed later in this section, both $\objective{\intensity{}}$ and $\objective{d}$ can be interpreted as cross-correlation scores between different parts of the present and past radar data.

\subsection{Direct intensity objective function}
\label{sec:direct_objective}

This objective function is inspired by direct methods in traditional monocular state estimation.
The goal is to estimate the continuous pose of the sensor to register/align the last radar scan with past radar data.
In this work, we opted for a local map that is updated `on-the-fly' as a way to combine past radar scans into a single image representation of the sensor's environment with intensity in Cartesian coordinates (as opposed to raw radar data in polar coordinates).
The motivation for using a local map instead of solely the last radar scan is the higher robustness to outliers such as moving vehicles or multipath echoes.

\subsubsection{Direct registration}

Let us assume the availability of an image-like local map $\map$ of the sensor's environment expressed in the radar's reference frame at time $\aztime{1}$.
The rows and columns of $\map$ correspond to discrete Cartesian coordinates.
The function $\interp{\bilinear}(\map, x, y)$ allows querying the local map intensity for any $x$ and $y \in \reals$ with bilinear interpolation.

To compute the direct registration objective between the incoming radar scan and the local map $\map$, we need to correct for the Doppler effect and the motion of the radar as
\begin{equation}
    \label{eq:pol_to_cart}
    \begin{bmatrix}
        \tilde{x}_{nm} \\ \tilde{y}_{nm}
    \end{bmatrix} = \rot{\aztime{1}}{\aztime{n}}\left(\left( \range{nm} \pm \frac{\Delta r_n}{2}\right) \begin{bmatrix} \cos(\az{n}) \\ \sin(\az{n})\end{bmatrix} \right) + \pos{\aztime{1}}{\aztime{n}},
\end{equation}
with $\Delta r_n$ the Doppler-induced shift~\eqref{eq:delta_range} as a function of the body-centric velocity and the beam azimuth:
\begin{equation}
    \label{eq:doppler_shift}
    \Delta r_n = \beta \begin{bmatrix} \cos(\az{n}) \\ \sin(\az{n})\end{bmatrix}^\top \rot{\world}{\aztime{n}}^\top\vel{\world}{\aztime{n}}.
\end{equation}
The sign $\pm$ is defined by the chirp direction for the azimuth $\az{n}$.
Please note that this objective function accounts for the Doppler effect but does not require an \ac{fmcw} radar with a triangular pattern.
Once the radar scan is corrected using the current estimate of the trajectory, the objective function is simply the cross-correlation score,
\begin{equation}
    \objective{\intensity{}} = \sum_{n=1}^{N}\sum_{m=1}^{M} \intensity{nm}\interp{\bilinear}(\map, \tilde{x}_{nm}, \tilde{y}_{nm}),
\end{equation}
between the corrected scan and the local map.

\subsubsection{On-the-fly local map}
The proposed local map consists of an image-like representation of the past radar data using a per-pixel low-pass filter to update $\map$ with the last radar scan and the corresponding state estimate.
Concretely, given the convergence of the optimization problem~\eqref{eq:optimisation}, the radar scan is corrected to the first timestamp of the following scan in Cartesian coordinates (using~\eqref{eq:pol_to_cart} replacing $\rot{\aztime{1}}{\aztime{n}}$ and $\pos{\aztime{1}}{\aztime{n}}$ with $\rot{\aztime{N+1}}{\aztime{n}}$ and $\pos{\aztime{N+1}}{\aztime{n}}$, respectively).
The corrected radar data is converted into an image-structure $\tilde{\image{}}_{\map}$ with the help of bilinear interpolation.
Then the local map update is 
\begin{equation}
    \map \gets (1-\gamma)\map + \gamma\tilde{\image{}}_{\map},
\end{equation}
with $\gamma \in [0,1]$ a user-defined parameter ($\gamma =0.1$ in our implementation).
Note that after receiving the very first radar frame, the local map is directly initialized with $\tilde{\image{}}_{\map}$.

\subsection{Doppler-based objective function}
\label{sec:doppler_objective}

\begin{figure*}
    \centering
    \def\offset{-6.5cm}
    \def\vdist{4.45cm}
    \def\ypos{1.5cm}
    \begin{tikzpicture}
        \node[inner sep=0,outer sep=0]{\includegraphics[width=0.99\linewidth, clip, trim=0.3cm 0.2cm 0.2cm 0.2cm]{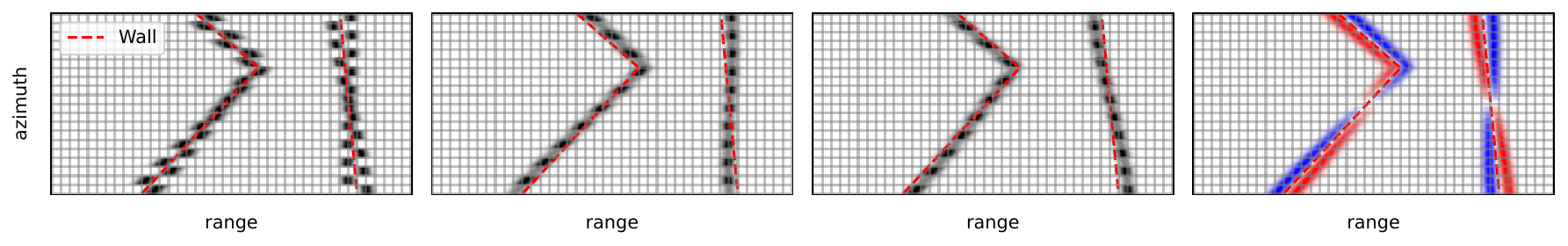}};
        \node[anchor=center] at (\offset,\ypos) {Raw radar data};
        \node[anchor=center] at (\offset+\vdist,\ypos) {GP-inferred image $\image{\up}$};
        \node[anchor=center] at (\offset+2*\vdist,\ypos) {GP-inferred image $\image{\down}$};
        \node[anchor=center] at (\offset+3*\vdist,\ypos) {$\image{\up} -\image{\down}$};
    \end{tikzpicture}
    \caption{Illustration of the proposed infilling. The raw Doppler distorted radar data (left) is split between up and down-chirp azimuths. Gaussian Process (GP) regression is used to interpolate every second azimuth (middle two, the interpolated portion is in grey). Both $\image{\up}$ and $\image{\down}$ virtually observe the same geometry without assuming that two consecutive azimuths of the raw data observe the same part of the environment (right, with blue positive and red negative).}
    \label{fig:gp_infill_intuition}
\end{figure*}
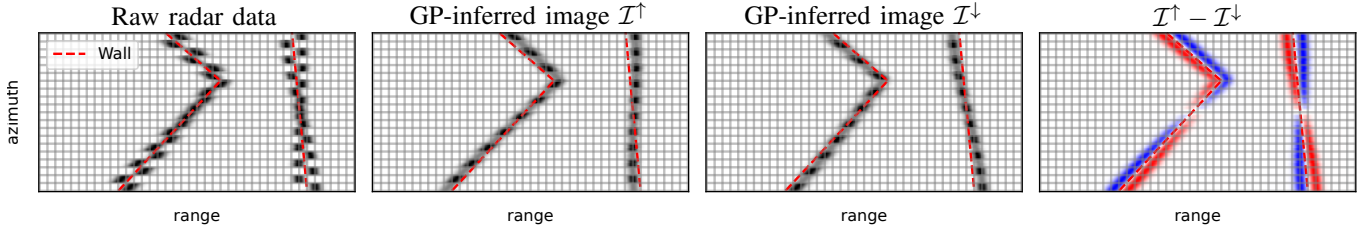

As presented in Section~\ref{sec:doppler_radar}, an \ac{fmcw} radar using a triangular modulation pattern (alternating up-chirp and down-chirp for each collected beam) allows for observability of the radial velocity along that beam if the consecutive beams observe the same part of the environment.
Unfortunately, this is not quite true in the context of mobile radar sensing, since the vehicle and the radar dish move between consecutively recorded returns. 
We propose to split the incoming radar scan into two image-like structures based on chirp direction and use \ac{gp} regression \eqref{eq:gp_infer} to `infill' the missing rows in both images.
This is illustrated in Fig.~\ref{fig:gp_infill_intuition}.
We denote the up-chirp and down-chirp infilled images as $\image{\up} = \{\az{n}, \range{m}, \intensityup{nm}\}$ and $\image{\down} = \{\az{n}, \range{m}, \intensitydown{nm}\}$, respectively.
Thanks to the \ac{gp} regression step, the part of the environment observed in each row of $\image{\up}$, be it observed or interpolated, is the same as in the corresponding row of $\image{\down}$.

\begin{figure}
    \centering
    \def\imgwidth{2.8cm}
    \def\hdist{0.05cm}
    \def\vdist{0.05cm}
    \def\legenddist{0.03cm}
    \def\clipsize{3}
    \begin{tikzpicture}
        \node [inner sep = 0em, outer sep = 0em] (staticraw) {\includegraphics[width=\imgwidth, clip, trim=\clipsize cm \clipsize cm \clipsize cm \clipsize cm]{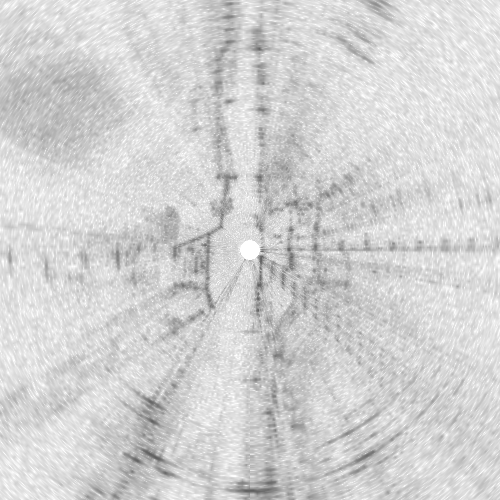}};
        \node [inner sep = 0em, outer sep = 0em, right=\hdist of staticraw] (staticup) {\includegraphics[width=\imgwidth, clip, trim=\clipsize cm \clipsize cm \clipsize cm \clipsize cm]{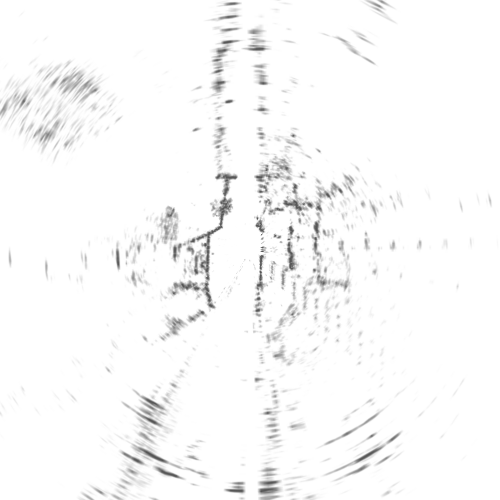}};
        \node [inner sep = 0em, outer sep = 0em, right=\hdist of staticup] (staticdiff) {\includegraphics[width=\imgwidth, clip, trim=\clipsize cm \clipsize cm \clipsize cm \clipsize cm]{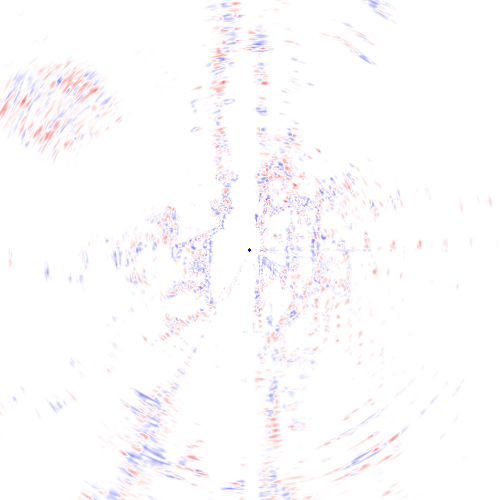}};
        \node [inner sep = 0em, outer sep = 0em, below=\vdist of staticraw] (movingraw) {\includegraphics[width=\imgwidth, clip, trim=\clipsize cm \clipsize cm \clipsize cm \clipsize cm]{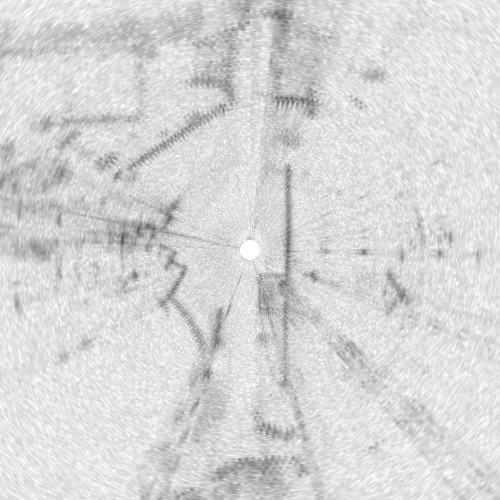}};
        \node [inner sep = 0em, outer sep = 0em, right=\hdist of movingraw] (movingup) {\includegraphics[width=\imgwidth, clip, trim=\clipsize cm \clipsize cm \clipsize cm \clipsize cm]{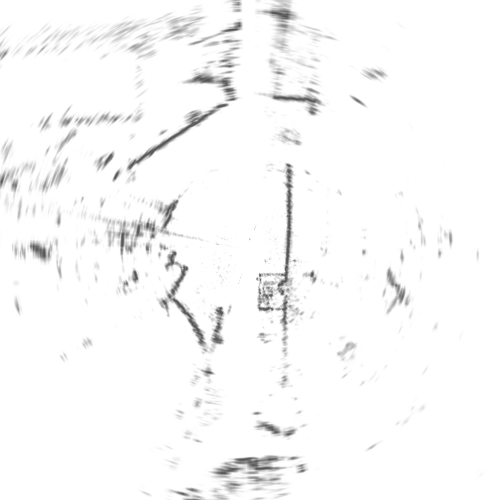}};
        \node [inner sep = 0em, outer sep = 0em, right=\hdist of movingup] (movingdiff) {\includegraphics[width=\imgwidth, clip, trim=\clipsize cm \clipsize cm \clipsize cm \clipsize cm]{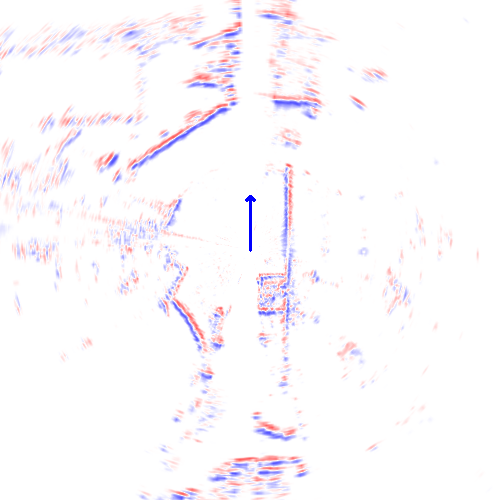}};
        \node[above=\legenddist of staticraw, inner sep = 0em, outer sep = 0em] {\scriptsize Raw radar scan};
        \node[above=\legenddist of staticup, inner sep = 0em, outer sep = 0em] {\scriptsize Up-image $\image{\up}$ (filtered)};
        \node[above=\legenddist of staticdiff, inner sep = 0em, outer sep = 0em] {\scriptsize Difference $\image{\up} - \image{\down}$};
        \node[left=\legenddist of staticraw, anchor=south, rotate=90, inner sep = 0em, outer sep = 0em] {\scriptsize Static};
        \node[left=\legenddist of movingraw, anchor=south, rotate=90, inner sep = 0em, outer sep = 0em] {\scriptsize Moving};
    \end{tikzpicture}
    \caption{Illustration of the GP-based infill highlighting the Doppler effect on radar data collected with a triangular frequency modulation pattern (the blue arrow is the sensor velocity).}
    \label{fig:gp_infill_real}
\end{figure}
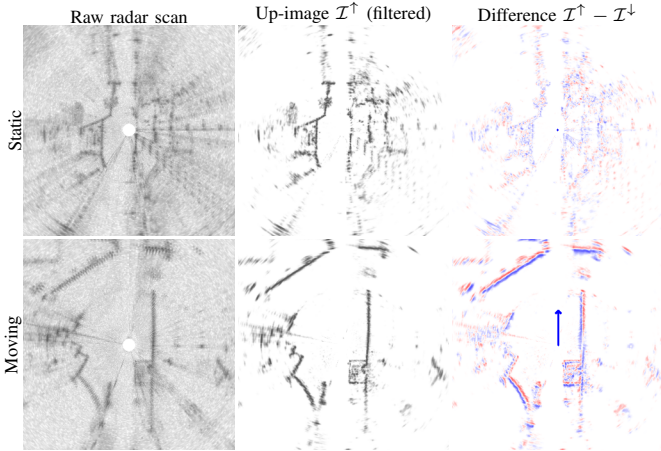
We show in Fig.~\ref{fig:gp_infill_real} that when the sensors are static, both $\image{\up}$ and $\image{\down}$ overlap well.
However, the Doppler shift is clearly visible in the difference between $\image{\up}$ and $\image{\down}$ when the sensors move.
The objective function $\objective{d}$ aims to constrain the estimated velocity $\vel{}{}(t)$ using $\image{\up}$ and $\image{\down}$.
First, we must compute the `range shift', with \eqref{eq:doppler_shift}, between each row of the two images as a function of the sensor velocity.
Then, we define the function $\interp{\linear}(\image{\up}, \az{n}, r)$ that queries the intensity value of $\image{\up}$ for any value of $r \in \reals$ using linear interpolation for each of the image's row. 
The objective function is a cross-correlation score between $\image{\down}$ and the Doppler-rectified version of $\image{\up}$:
\begin{equation}
    \objective{d} = \sum_{n=1}^{N}\sum_{m=1}^{M} \intensitydown{nm}\interp{\linear}(\image{\up}, \az{n}, \range{m}+\Delta r_n).
\end{equation}
The objective function $\objective{d}$ is maximised when the body-centric velocity estimate $\rot{}{}(\aztime{n})^\top\vel{}{}(\aztime{n})$ corresponds to the actual velocity of the sensor.

\subsection{Motion models}
\label{sec:motion_models}

In this section, we present different ways to obtain the continuous trajectory functions $\rot{\world}{t}$, $\pos{\world}{t}$, and $\vel{\world}{t}$ from the state variables $\state$.
As $\rot{\world}{t}$ is in $SO(2)$, the rotation can easily be represented with a simple rotation angle $\theta(t)$ as
\begin{equation}
    \rot{\world}{t} = \rot{\world}{\aztime{1}} \begin{bmatrix}
        \cos(\theta(t)) & \sin(\theta(t))
        \\
        -\sin(\theta(t)) & \cos(\theta(t))
    \end{bmatrix},
\end{equation}
with $\rot{\world}{\aztime{1}}$ being the starting rotation obtained from the state estimate of the previous radar scan.
We introduce two different models to obtain $\theta(t)$ in Subsections~\ref{sec:const_ang_vel}, and~\ref{sec:gyro}.
Regarding the translation and velocity, we use a simple constant body-centric velocity model in Subsection~\ref{sec:const_body_vel}.
For convenience, we define $\state = \{\state_{\rot{}{}}, \vel{b}{}\}$ with $\state_{\rot{}{}}$ the rotational state variables, and $\vel{b}{}$ the body-centric velocity.

\subsubsection{Constant angular velocity}
\label{sec:const_ang_vel}

In the absence of a gyroscope, the proposed method can rely on the assumption of constant angular velocity $\omega$ for the duration of any radar scan with $\theta(t) = \omega (t-\aztime{1})$.
Therefore, the state variable is $\state_{\rot{}{}} = \{\omega\}$.

\subsubsection{Gyro preintegration}
\label{sec:gyro}

Inspired by \cite{legentil2024truly}, when using a gyroscope, the change of orientation of the sensor suite is defined as a function of the angular rate measurements.
To account for the frequency discrepancy between the radar azimuths and the gyroscope measurements, it leverages continuous preintegration concepts drawn from \cite{legentil2020gpm} and \cite{legentil2023latent}.
For the sake of simplicity and lightweight computation, we use a simple piece-wise linear model
\begin{equation}
    \label{eq:continuous_integration}
    \theta(t) = \theta(\imutime{i}) + \frac{(\angvel{i+1}+\angvel{i})(t-\imutime{i})}{2(\imutime{i+1}-\imutime{i})} + \angvel{i}(t-\imutime{i}),
\end{equation}
with $\angvel{i} = \gyr{i} - b$, $t\in[\imutime{i}, \imutime{i+1}]$, 
We decided not to include the gyroscope bias $b$ in the estimated state.
In our experiments, biases range from $\approx5e^{-5}$ to $\approx4e^{-3}$ corresponding to a maximum of $0.057^\circ$ drift over the $250\,\si{\ms}$ of a scan.
With the typical $0.9^\circ$ radar angular resolution, the signal-to-noise ratio is too high to enable robust bias estimation in a `frame-to-frame` approach like ours. 
This is especially true as the proposed objective functions are not probabilistic, thus impeding the integration of additional residuals/objective functions from other modalities and the probabilistic prior/previous knowledge of the bias.
Accordingly, when using a gyroscope, the rotational state is empty $\state_{\rot{}{}} = \varnothing$.
However, we use a separate online bias estimation strategy based on a simple yet effective heuristic: when the vehicle's velocity is zero, so is the angular rate.
Accordingly, the bias estimate is initialized by averaging the first $Q$ raw gyroscope measurements collected when the velocity estimate is under $5\,\si{\cm/\s}$.
Then it is updated with a low-pass filter whenever the velocity estimate is under $5 \, \si{\cm/\second}$.

\subsubsection{Constant body-centric velocity}
\label{sec:const_body_vel}

This model assumes a constant body-centric velocity $\vel{b}{} \in \reals^2$ during any radar scan.
The velocity in the global frame is $\vel{\world}{t} = \rot{\world}{t}\vel{b}{}$.
The position is 
\begin{equation}
    \label{eq:pos_const_vel}
    \pos{\world}{t} = \pos{\world}{\aztime{1}} + \left(\int_\imutime{1}^t \rot{\world}{s}ds\right)\vel{b}{},
\end{equation}
with $\pos{\world}{\aztime{1}}$ the position resulting from the state estimation during the previous radar scan.
The integral in~\eqref{eq:pos_const_vel} is computed analytically when using the constant angular velocity assumption.
Otherwise, when using a gyroscope, it is numerically preintegrated using a piece-wise linear model similar to \eqref{eq:continuous_integration} upon each of the matrix elements.

\section{Experiments}

\subsection{Implementation}

The proposed method has been implemented in Python using PyTorch for matrix and algebra operations.
Accordingly, it can easily be run on either a GPU or a CPU.
Note that we have implemented~\eqref{eq:optimisation} with analytical Jacobians and our solver without relying on PyTorch's automatic differentiation and integrated solvers.
When using a gyroscope and the Doppler-based velocity constraints, our method converges in 13.5 iterations on average.

\subsubsection{Gaussian process regression}

\ac{gp} regression is known to suffer cubic computational complexity due to the matrix inversion in~\eqref{eq:gp_infer}.
With around 2 million points in each radar scan, naive \ac{gp} regression to generate $\image{\up}$ and $\image{\down}$ is intractable.
Fortunately, efficient computation strategies are enabled by the grid-pattern (image-like) nature of both the radar data and the inference locations in the proposed \ac{gp}-based infill step.
By only considering the data in a $U\times V$ neighbourhood around each inference location, $\kernelvec{}{\abscissavec^*}{\abscissamat} \left(\kernelmat{}{\abscissamat}{\abscissamat} + \sigma^2\identitymat\right)^{-1}$ can be precomputed and the multiplication with $\ordinatevec$ is a simple convolution operation.
Accordingly, with a GPU implementation, the inference of $\image{\up}$ and $\image{\down}$ is extremely efficient.

Similarly to \cite{lisus2025doppler}, each row of the \ac{gp}-inferred image is independently filtered to attenuate multipath and specular noise.
First, the intensity standard deviation is computed and any range bin with a value inferior to twice the deviation is nullified.
Then the intensity values are normalized so that the maximum value equals one.
Per-azimuth Gaussian blur is applied to smooth the discontinuities introduced in the previous steps.
Finally, each intensity value is cubed to augment contrast.

\subsubsection{Optimisation}
The optimization problem~\eqref{eq:optimisation} is solved with a gradient ascent algorithm with an update step  
\begin{equation}
    \state \gets \state + \step \frac{\nabla(\objective{d}+\objective{\intensity{}})}{\lVert\nabla(\objective{d}+\objective{\intensity{}})\rVert},
\end{equation}
where $\nabla$ is the gradient operator, and $\step$ is a constant equal to 0.1 at the start of the optimization and halved for every non-ascending step.
Note that the objective functions are element-wise products of intensity values.
Thus, the computations linked to intensity values that are equal to zero do not contribute to the optimization problem and can be omitted to lower the computational burden.

\subsubsection{Robust weighting}

\begin{figure}
    \centering
    \def\legenddist{0.1cm}
    \def\legendtextsize{\small}
    \def\hdist{0.1cm}
    \def\vdist{0.8cm}
    \def\imgscale{0.49}
    \begin{tikzpicture}
        \node[inner sep=0, outer sep=0](raw){\includegraphics[width=\imgscale\columnwidth]{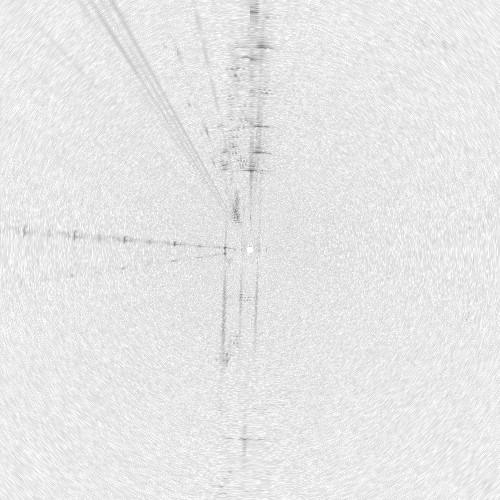}};
        \node[inner sep=0, outer sep=0, right=\hdist of raw](diff){\includegraphics[width=\imgscale\columnwidth]{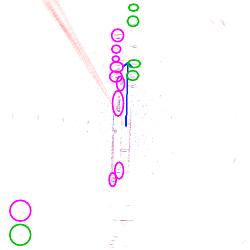}};
        \node[inner sep=0, outer sep=0, above=\legenddist of raw]{\legendtextsize Raw radar scan};
        \node[inner sep=0, outer sep=0, above=\legenddist of diff]{\legendtextsize Difference $\image{\up} - \image{\down}$};
        \node[anchor=west] at (2.85,-1.51){\scriptsize Vehicles in the opposite lane};
        \node[anchor=west] at (2.85,-1.91){\scriptsize Vehicles in the same lane};
        \draw[<->] (1.4, -2) -- (1.4,2);
        \node[anchor=west] at (1.4,0){\scriptsize 400m};

        \node[inner sep=0, outer sep=0, below=\vdist of raw](cam){\includegraphics[width=\imgscale\columnwidth]{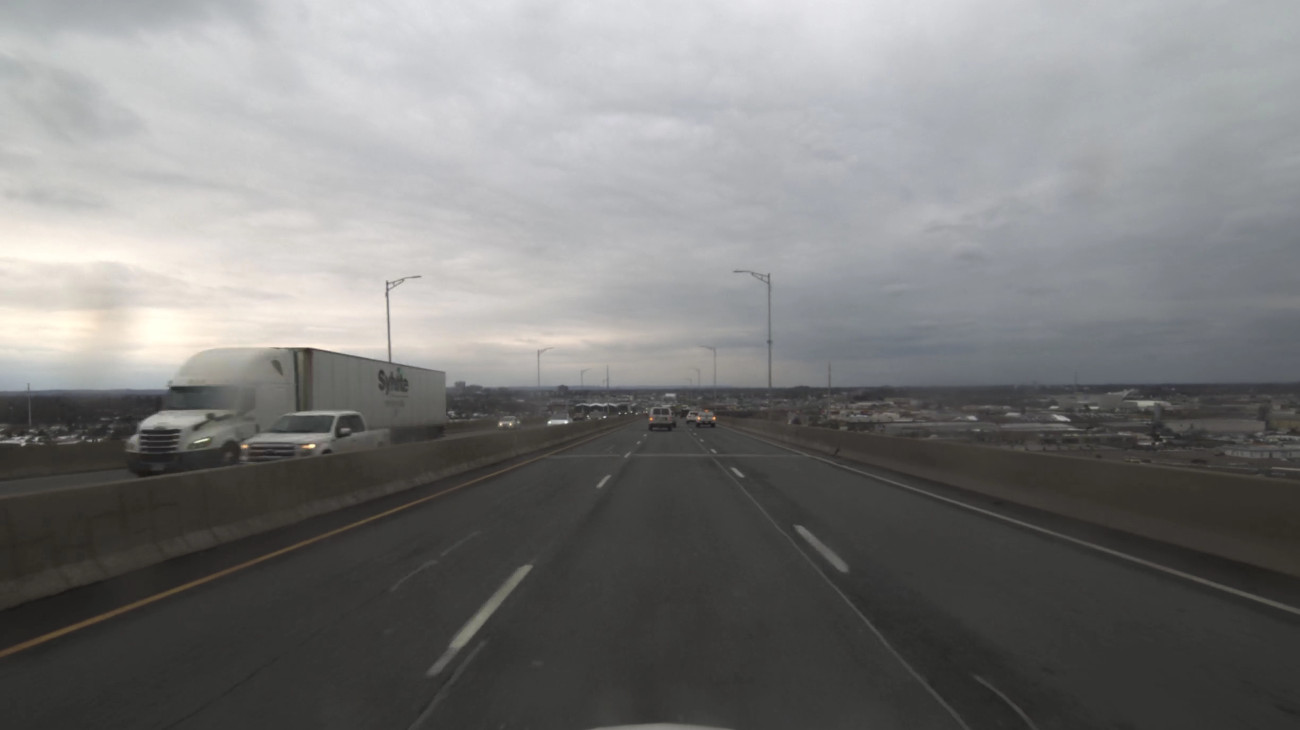}};
        \node[inner sep=0, outer sep=0, below=\vdist of diff](skyway){\includegraphics[width=\imgscale\columnwidth]{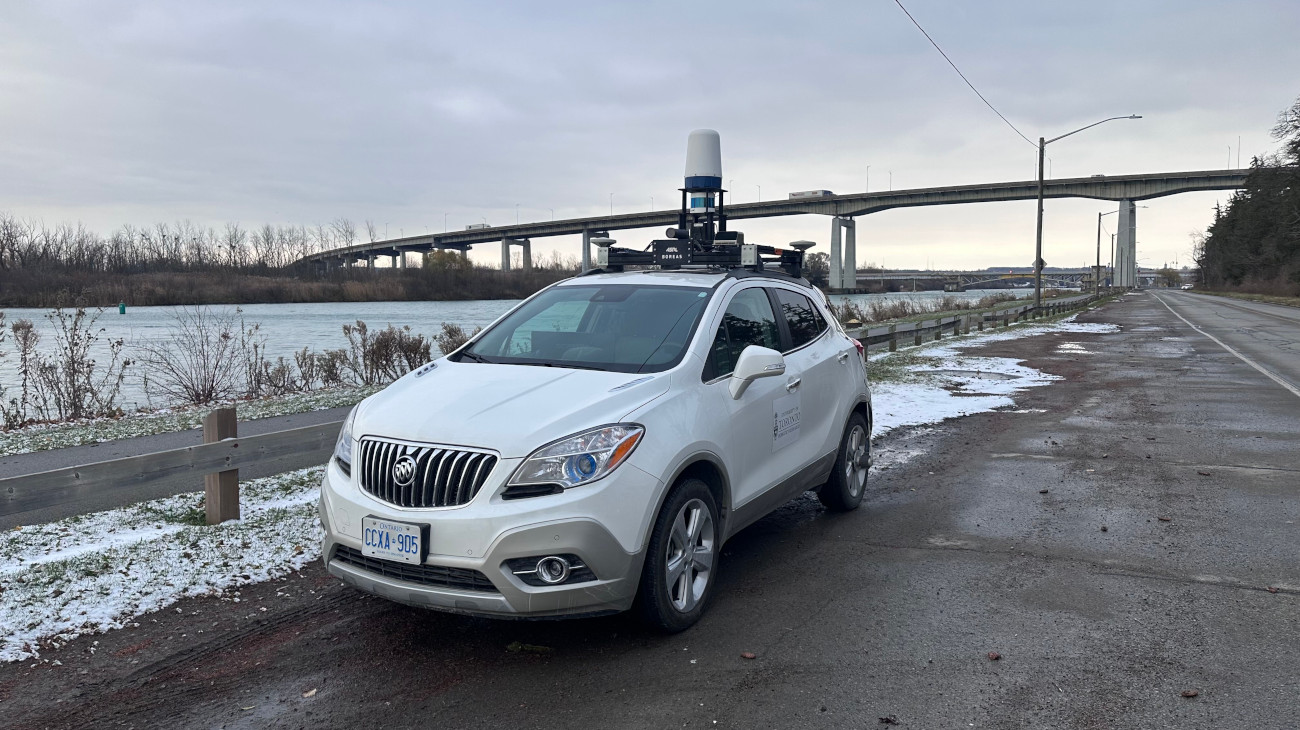}};
        \node[inner sep=0, outer sep=0, above=\legenddist of cam]{\legendtextsize On-board camera view};
        \node[inner sep=0, outer sep=0, above=\legenddist of skyway]{\legendtextsize Skyway ground view};
    \end{tikzpicture}
    \caption{Illustration of a challenging situation with a line of vehicles coming the other way while driving over a skyway (very few static elements visible in the radar data).}
    \label{fig:challenging}
\end{figure}
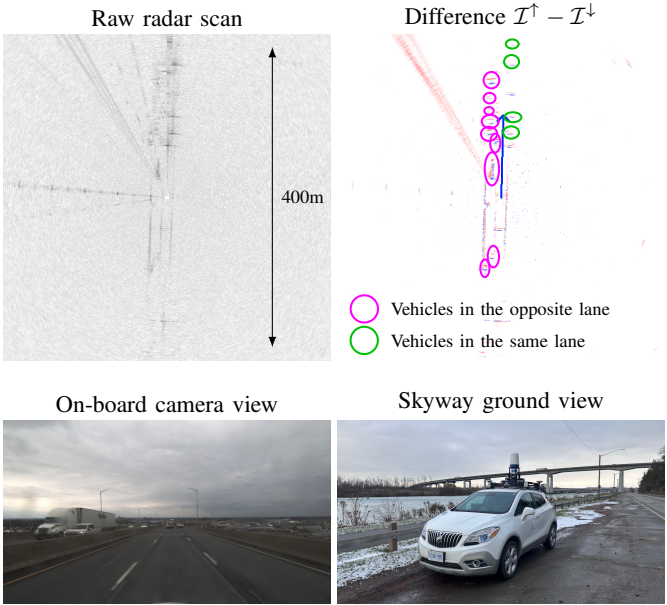

The proposed objective functions $\objective{d}$ and $\objective{\intensity{}}$ are generally robust to outliers such as moving vehicles and specular noise.
However, on very rare occasions (2 frames among 42k in our automotive experiments), a very high proportion of consistent outliers can lead to erroneous state estimates. 
A typical example is illustrated in Fig.~\ref{fig:challenging} when the sensing platform is moving over a bridge/skyway with a very small amount of static reflectors in the surroundings, and a line of vehicles comes the other way.
It can make the velocity estimate suddenly jump unrealistically.
Such scenarios are easily detected with a simple threshold on the sensor's acceleration from one scan estimate to the next.
When detected, we re-run the optimization weighting the objective functions $\objective{\intensity{}}$ and $\objective{d}$ as
\begin{equation}
    \begin{aligned}
    \objective{\bullet} &= \sum_{n=1}^{N}\sum_{m=1}^{M}\rho^\bullet_{nm}\intensity{nm}\interp{\cdot}(\cdot, \cdot, \cdot),
    \\
    \rho^\bullet_{nm} &= \left(\lvert \intensity{nm} - \interp{\cdot}(\cdot,\cdot,\cdot) \rvert - 1 \right)^6,
    \end{aligned}
\end{equation}
with $\bullet = d$ or $\intensity{}$.
Intuitively, the weighting $\rho$ acts as an outlier-rejection mechanism by down-weighting the objective components that are too dissimilar in terms of intensity.

\subsubsection{Velocity bias in Doppler constraint}
\label{sec:vy_bias}
Similarly to \cite{lisus2025doppler}, we have empirically observed a velocity bias when performing Doppler-only velocity estimation.
This bias mostly concerns the lateral velocity of the sensing platform.
We believe this bias is partly explained by the disparity between the actual measurement process and the simple measurement model that assumes the information contained in a row of the radar data corresponds to a single azimuth/timestamp: in reality, the radar's scanning pattern is continuous.
Thus, every measurement from a row could correspond to a different azimuth and timestamp.
We think that the motion model's `inaccuracy' leads to a velocity bias that appears as environment-dependent due to the ignored correlation between the data range and azimuth/timestamp.
The authors of \cite{lisus2025doppler} addressed this issue by fitting a linear model to the velocity error using held-out data with accurate velocity ground-truth.
Then, the linear model is used to correct the velocity estimates directly.
In this work, we propose to leverage the non-slip kinematics constraint to estimate and compensate for the lateral component of the velocity bias online without a `calibration' phase.

Using the angular rate and the extrinsic calibration between the radar and the vehicle's rear axle, the velocity estimated based solely on the Doppler-based constraint $\objective{d}$ is projected into the axle reference frame.
The $y$ component is projected back to the radar frame to update a low-pass filter that tracks the lateral velocity bias.
The filter's output is then used to correct the radial velocities when computing the shifts~\eqref{eq:doppler_shift} for Doppler distortion correction.
This estimation and correction process is only applied when $\objective{d}$ is part of the optimization problem~\eqref{eq:optimisation}.
It is important to note that this mechanism does not strictly enforce the non-slip constraint thanks to the low-pass filter.
Accordingly, the vehicle can be subject to momentary side slips without altering the state estimates.

\subsection{Metrics}

To benchmark our algorithm, we use well-established metrics in the odometry literature.
For datasets with ground-truth orientation, the KITTI relative translation and rotation errors are adopted.
Succinctly, for one pose estimate, the estimated trajectory is aligned with the ground-truth.
Then, after a certain distance travelled from the aligned poses (segments varying from 100 to 800$\,\si{\m}$), the estimated and ground-truth poses are used to compute position and orientation errors relative to the segment lengths.
This process is repeated every 5 poses of the estimated trajectory, and the results are reported as the average relative errors in \% and $\si{\degree}/100\,\si{\m}$.
In the absence of ground-truth orientation, we adapted the \ac{rpe} introduced in \cite{zhang2018evaluation} to SE(2) trajectories.
It corresponds to the position \ac{rmse} between aligned segments of the ground-truth and the estimated trajectory.
The length of segments [50\,\si{m}, 100\,\si{m}, 150\,\si{m}, 200\,\si{m}].
Similarly to the KITTI metric, we display the \ac{rmse} error as a percentage of the segment lengths.

\subsection{Doppler-enabled automotive odometry}
\label{sec:exp_automotive}

In this section, we provide a detailed analysis of our method's performance on a dataset collected with our automotive sensing platform.
Note that the radar used in this set of experiments uses a triangular frequency modulation pattern, thus enabling the use of the Doppler-based velocity constraints $\objective{d}$ in \eqref{eq:optimisation}.

\subsubsection{Dataset description}

The goal of this set-up is to replicate experiments from~\cite{lisus2025doppler}.
To do so, we have collected data with an automotive platform equipped with a Navtech RAS6 radar, a Silicon Sensing DMU41 \ac{imu}, a Velodyne Alpha Prime lidar, and an accurate Applanix RTK-GNSS/INS solution for ground-truth with post-processing.
All the sensors have been accurately synchronized.
The data used for benchmarking has been collected repeatedly four times along four routes (16 sequences in total) displaying an increasing level of difficulty: \emph{Suburbs}, \emph{Highway}, \emph{Tunnel}, and \emph{Skyway}.
The first two environments provide a relatively large amount of geometric constraints via the presence of many buildings and other human-made structures in the vehicle's surroundings.
In \emph{Tunnel} and \emph{Skyway}, the geometric features are either degenerate (not constraining all the movement's axes) or close to nonexistent, respectively.
In these scenarios, only Doppler-induced information can lead to usable odometry performances.
The skyway scenario is especially challenging as for a couple of kilometres the radar does not observe any significant static features except for sparse lamp posts and weak returns from the side barriers (the radar is mounted on the top of the vehicle and the barriers are relatively short, thus the bulk of the emission cone passes over the barriers).
Accordingly, the ratio of moving to static elements in the radar's \ac{fov} is very high.
The length and velocity characteristics of each sequence type are given in Table \ref{tab:doppler} along with the results discussed later in Section~\ref{sec:doppler_results}.
Fig.~\ref{fig:teaser} (left) shows our collection vehicle while recording a \emph{Tunnel} sequence.

\subsubsection{Baselines}
To benchmark the proposed algorithm, we have replicated four baselines from~\cite{are_we_ready_for} and~\cite{lisus2025doppler}, three based on radar data and one on lidar.
The first baseline, denoted CT-R, corresponds to the odometry component of the publicly available `teach and repeat' framework~\cite{are_we_ready_for}.
It consists of an ICP-based continuous-time scan registration using point clouds extracted from the raw radar data.
The trajectory estimates are computed via a sliding window optimization based on a continuous-time \ac{gp} state representation.
Note that the velocity estimates are used to compensate for the Doppler-induced shift of the radar points.
The second baseline is DG and replicates the work from~\cite{lisus2025doppler} on per-azimuth radial velocity extraction, followed by a RANSAC-based outlier rejection and robust optimization.
With the constant-velocity assumption, it allows for the estimation of the vehicle's velocity for each radar scan.
Combined with the gyroscope's angular rate, it enables association-free radar odometry, even in geometrically feature-deprived environments.
Despite not extracting any point cloud out of the radar data, one can consider the radial velocity as `features' extracted in a `front-end' before an optimization-based state estimation step.
Third is CT-RDG, the integration of the per-scan velocity estimates from~\cite{lisus2025doppler} and the gyroscope measurements within CT-R.
Finally, the lidar baseline denoted CT-LG consists of the continuous-time ICP from~\cite{are_we_ready_for} with additional gyroscope constraints.
The various parameters of the baseline, as well as the radar-specific value of $\beta$, have been tuned using an extra sequence of each type.

\subsubsection{Results}
\label{sec:doppler_results}

\begin{table*}
    \centering
    \caption{Average relative pose accuracy of the proposed method and several baselines on our dataset (best radar-based method in bold).}
    \setlength{\tabcolsep}{2pt}
    \begin{tabularx}{\linewidth}{lYYYYYYY}
        \toprule
        \textbf{Sequence type} \scriptsize(length, avg. / max. vel.) & \textbf{CT-R} \cite{are_we_ready_for} & \textbf{DG} \cite{lisus2025doppler} & \textbf{CT-RDG} \cite{lisus2025doppler} & \textbf{CT-LG} \cite{are_we_ready_for} & \textbf{DRO-D} \scriptsize(ours) & \textbf{DRO-G} \scriptsize(ours) & \textbf{DRO-GD} \scriptsize(ours)
        \\
        \midrule
        Suburbs \scriptsize($4\times7.9\, \si{\km}$, $8.1$ / $18.6\,\si{\m/\s}$) & 1.22 / 0.37 & 0.60 / 0.04 & 0.32 / 0.04 & 0.15 / 0.07 & 0.72 / \textbf{0.02} & \textbf{0.18} / \textbf{0.02} & 0.19 / \textbf{0.02}
        \\ 
        Highway \scriptsize($4\times9.3\, \si{\km}$, $11.1$ / $27.0\,\si{\m/\s}$) & 1.92 / 0.47 & 0.74 / 0.04 & 0.49 / 0.04 & 0.19 / 0.06 & 0.84 / \textbf{0.02} & \textbf{0.16} / \textbf{0.02} & 0.24 / \textbf{0.02}
        \\
        Tunnel \scriptsize($4\times 1.9\, \si{\km}$, $8.9$ / $28.2\,\si{\m/\s}$) & 35.90 / 0.96 & 1.35 / 0.04 & 0.70 / 0.04 & 3.35 / 0.11 & 0.54 / \textbf{0.02} & 7.12 / \textbf{0.02} & \textbf{0.34} / \textbf{0.02}
        \\
        Skyway \scriptsize($4\times 11.1\, \si{\km}$, $18.2$ / $31.7\,\si{\m/\s}$) & 35.24 / 1.77 & 1.17 / 0.02 & 0.86 / 0.02 &  16.98 / 0.24 & 0.71 / \textbf{0.01} & 45.76 / \textbf{0.01} & \textbf{0.40} / \textbf{0.01}
        \\
        \bottomrule
        & \multicolumn{7}{c}{\scriptsize KITTI odometry metric reported as \textit{XX / YY} with \textit{XX} [\%] and \textit{YY} [$\si{\degree}/100\,\si{\m}$] the translation and orientation errors, respectively.}
    \end{tabularx}
    \label{tab:doppler}
\end{table*}

Table~\ref{tab:doppler} shows the average odometry error (KITTI metrics) obtained with each baseline and our algorithm in the four environments.
While a thorough ablation study is conducted in the following subsection, here, we consider three variations of DRO that all rely on the gyroscope integration for the orientation estimation.
The first variant, denoted DRO-D, solely uses the Doppler-based velocity constraint $\objective{d}$ to estimate the linear dynamics of the system.
The second one, DRO-G, uses solely the direct registration objective function $\objective{\intensity{}}$.
Note that DRO-G still leverages Doppler compensation of the radar data as part of $\objective{\intensity{}}$.
Ultimately, DRO-GD combines both $\objective{d}$ and $\objective{\intensity{}}$ as presented in~\eqref{eq:optimisation}.

Overall, DRO-GD significantly outperforms every other radar methods.
While the performance gap between DRO-GD and DRO-G is minimal in the well-structured suburban environment, the difference increases as the data becomes more challenging.
The best radar baseline, CT-RDG, shows errors around twice as large as ours throughout all the sequences
When considering `Doppler-only' approaches, DG~\cite{lisus2025doppler} and DRO-D perform similarly in \emph{Suburbs} and \emph{Highway}.
In the more challenging environments, DRO-D displays a noticeable advantage.
All these observations suggest that the direct nature of the proposed approach (considering all the radar information in a single optimization without any feature extraction) and our novel \ac{gp}-based infill for Doppler data provide a greater level of robustness.
Without the Doppler-based velocity constraint, DRO-G is not able to reliably perform odometry in challenging scenarios, similar to CT-R.
Despite not completely failing on all the \emph{Tunnel} sequences (3 out of 4 runs present a translation error under 2.5\%), the Doppler-compensation that happens in $\objective{\intensity{}}$ is not sufficient for robust estimation in feature-deprived environments.
Only the lidar baseline outperforms DRO-G in some feature-dense environments (\emph{Suburbs}).
However, CT-LG poorly performs in feature-deprived sequences (\emph{Tunnel} and \emph{Skyway}) while DRO-GD keeps a similar level of performance.

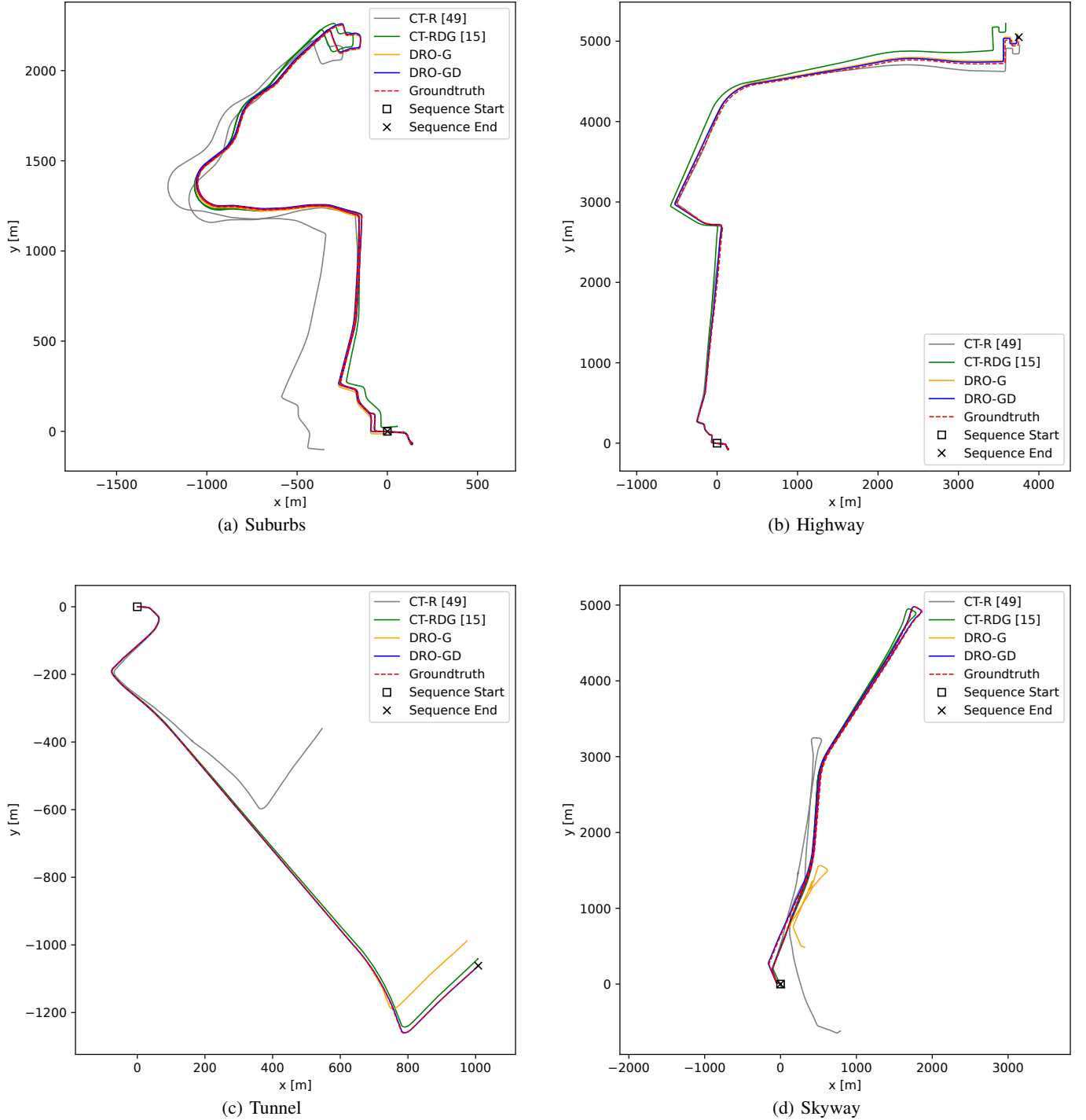
\begin{figure*}
    \centering
    \def\subfigdist{0.0cm}
    \def\vdist{1cm}
    \def\hdist{0.5cm}
    \def\imgscale{0.49}
    \def\subfigtextsize{\small}
    \begin{tikzpicture}
        \node[inner sep=0, outer sep=0] (suburbs) {\includegraphics[clip, width=\imgscale\linewidth]{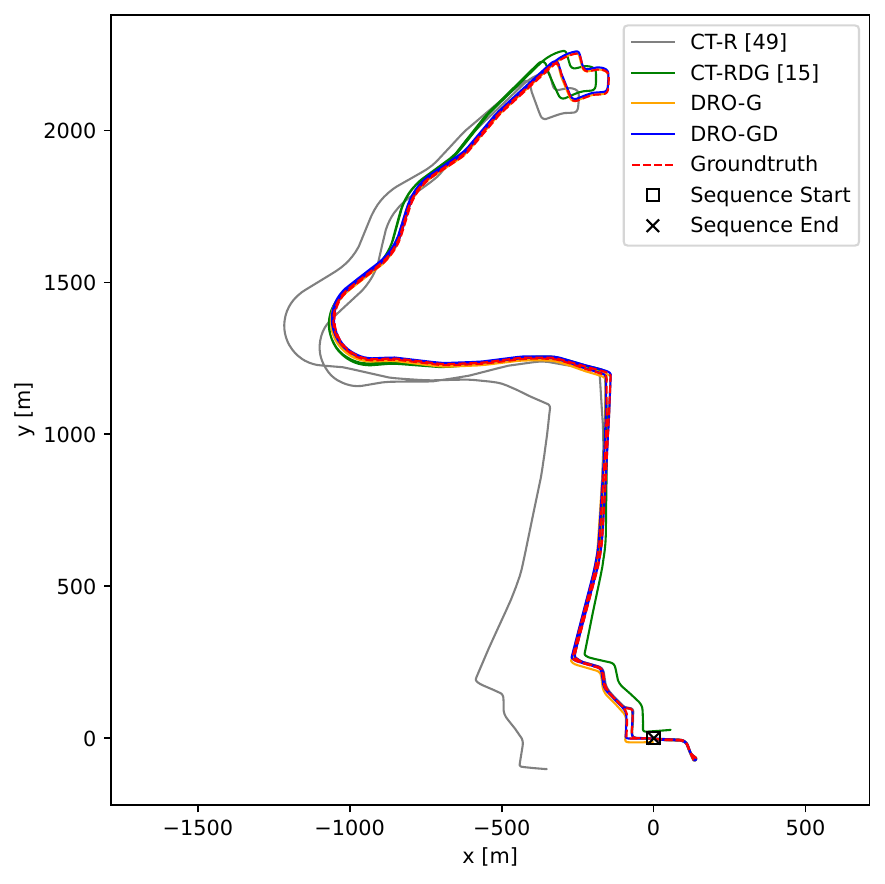}};
        \node[inner sep=0, outer sep=0, right=\hdist of suburbs] (highway) {\includegraphics[clip, width=\imgscale\linewidth]{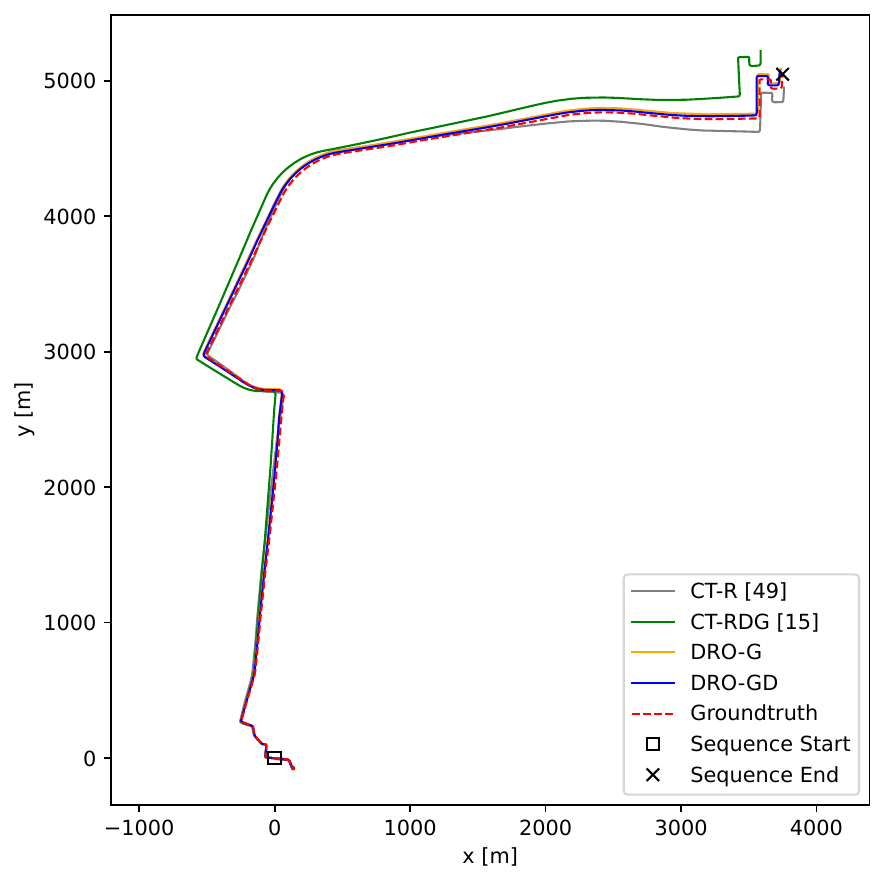}};
        \node[inner sep=0, outer sep=0, below=\vdist of suburbs] (tunnel) {\includegraphics[clip, width=\imgscale\linewidth]{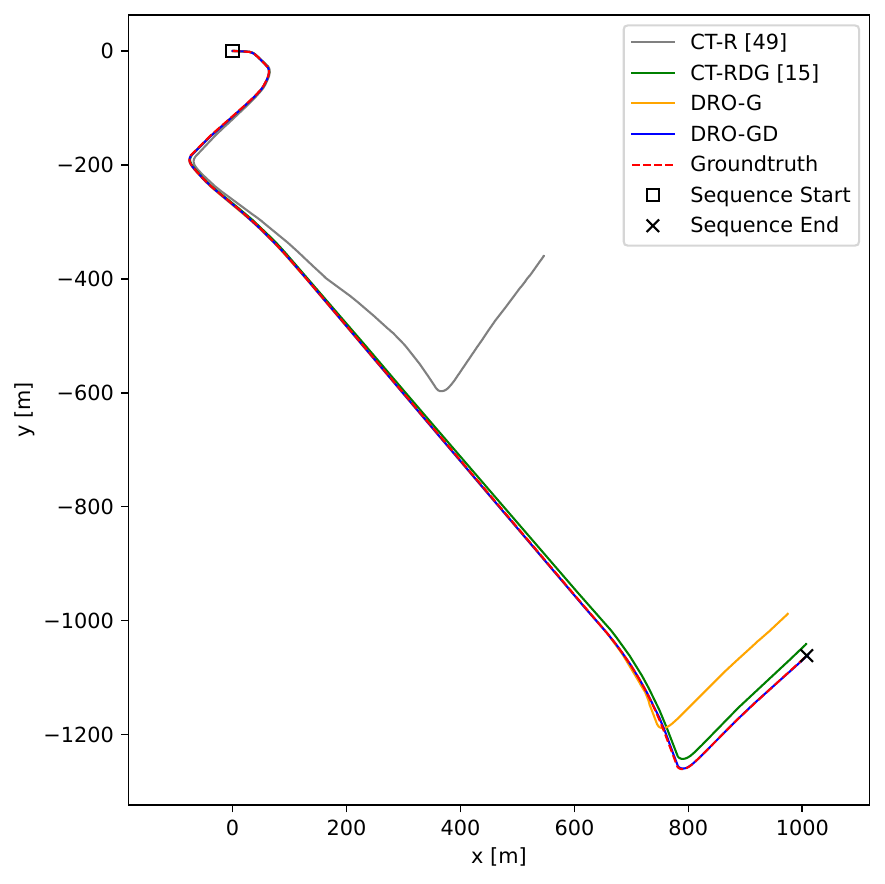}};
        \node[inner sep=0, outer sep=0, below=\vdist of highway] (skyway) {\includegraphics[clip, width=\imgscale\linewidth]{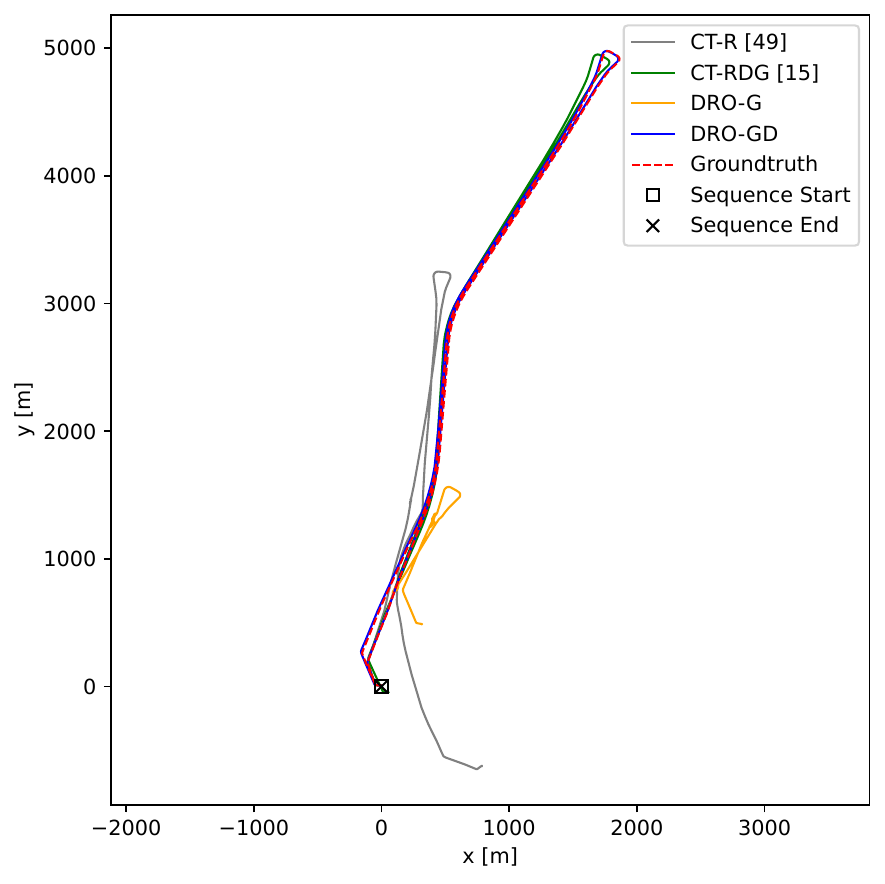}};
        
        \node[inner sep=0, outer sep=0, below=\subfigdist of suburbs] {\subfigtextsize (a) Suburbs};
        \node[inner sep=0, outer sep=0, below=\subfigdist of highway] {\subfigtextsize (b) Highway};
        \node[inner sep=0, outer sep=0, below=\subfigdist of tunnel] {\subfigtextsize (c) Tunnel};
        \node[inner sep=0, outer sep=0, below=\subfigdist of skyway] {\subfigtextsize (d) Skyway};
    \end{tikzpicture}
    \caption{Trajectory estimate samples for the different sequence types.}
    \label{fig:trajectories}
\end{figure*}

Visualizations of the raw data, velocity estimate and local map are provided in the supplementary materials\footnote{\url{https://youtu.be/QYVYUbNziwY}}. Fig.~\ref{fig:trajectories} provides a sample of the estimated trajectories for each sequence type.
It is interesting to note that despite relatively similar KITTI metrics, CT-RDG and DRO-DG result in fairly different trajectories, with DRO-DG being much closer to the ground-truth.
Part of the explanation is that the KITTI metrics only consider segments up to $800\,\si{m}$ where the small orientation drift of CT-RDG has a weak impact on the position at the end of each segment.
However, along the full trajectory length, the orientation drift leads to large position errors that are visually noticeable.
This observation supports the fact that for kilometre-long odometry, the direct integration of the gyroscope data can provide better orientation estimates than \ac{sota} odometry.

\subsubsection{Ablation study}

We have conducted a thorough ablation study to determine which of the presented mechanisms impacts the algorithm's accuracy the most.
Due to the extremely challenging nature of the Skyway sequences, we have chosen to leave them out of this study, as other factors, such as the high proportion of outliers, can have a greater impact on the results.
Table~\eqref{tab:ablation} shows the relative pose error of the proposed DRO-GD when various elements are removed from the pipeline.
For the `no local map' row, DRO-GD is run replacing the proposed local map updated on the fly with simply the last motion and Doppler-corrected scan.
As expected, removing the gyroscope\footnote{For the gyroscope-less operation, we correct the estimated state with an angular velocity bias calibrated using held-off data ($0.034\si{\degree/s}$ for \emph{Suburbs} and \emph{Highway}, and $0.057\si{\degree/s}$ for \emph{Tunnel}).
We have also applied the robust weighting when the change in orientation between consecutive scans is not realistic. Without the bias correction, we obtain relative translation errors of $1.37\%$, $1.61\%$, and $3.67\%$ for \emph{Suburbs}, \emph{Highway}, and \emph{Tunnel}, respectively.} corresponds to the largest performance degradation, leading to accuracy slightly better than CT-R in Table~\ref{tab:doppler} (also without a gyroscope) for feature-dense environments.
Estimating the system's orientation in a continuous manner using only radar data is challenging due to the low angular resolution of the sensor ($0.9\si{\degree}$) and the beam width ($1.8\si{\degree}$), even when leveraging a motion prior like CT-R.
Interestingly, in more challenging scenarios like the \emph{Tunnel} sequences, the drop in our method's accuracy is not as large as for CT-R thanks to the proposed Doppler-based velocity constraint.
Regardless, the velocity \ac{rmse} is not significantly impacted by the absence of a gyroscope.
As for the other variations, the difference with DRO-GD is limited, and on average, our algorithm still outperforms the best radar baseline CT-RDG.

\begin{table}
    \centering
    \caption{Pose accuracy ablation study: each row corresponds to a marginal difference with DRO-GD.}
    \begin{tabularx}{\columnwidth}{lccc|Y}
        \toprule
        \textbf{Method} & \textbf{Suburbs} & \textbf{Highway} & \textbf{Tunnel} & \textbf{Vel. er.}
        \\
        \midrule
        DRO-GD \tiny(ours) & \textbf{0.19} / \textbf{0.02} & \textbf{0.24} / \textbf{0.02} & \textbf{0.34} / \textbf{0.02} & 0.119
        \\ 
        No gyr. bias & 0.27 / 0.03 & 0.31 / 0.04 & 0.42 / 0.04 & 0.119
        \\
        No vel. bias & 0.22 / 0.02 & 0.29 / 0.02 & 0.31 / 0.02 & 0.120
        \\
        No local map & 0.35 / 0.02 & 0.43 / 0.02 & 0.40 / 0.02 & \textbf{0.107}
        \\
        No gyroscope & 1.00 / 0.24 & 1.56 / 0.38 & 2.92 / 0.99 & 0.132
        \\
        \bottomrule
        \multicolumn{4}{c|}
        {\shortstack[l]{\rule{0pt}{2.ex}\scriptsize KITTI odometry metric reported as \textit{XX / YY} with \textit{XX} [\%] \\\scriptsize and \textit{YY} [$\si{\degree}/100\,\si{\m}$] the trans. and rot. errors, respectively.}} & 
        \multicolumn{1}{c}{\shortstack{\scriptsize Avg RMSE\\\scriptsize$[$\si{\m/\s}$]$}} 
    \end{tabularx}
    \label{tab:ablation}
\end{table}

\begin{figure}
    \centering
    \includegraphics[width=0.99\columnwidth]{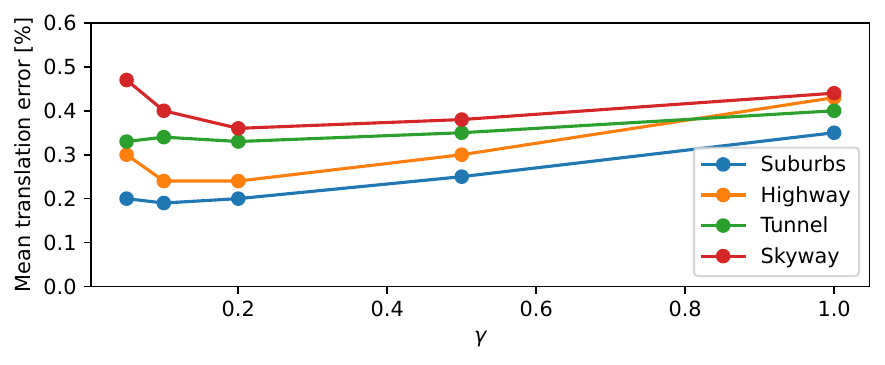}
    \vspace{-0.9cm}
    \caption{Sensitivity analysis of DRO-GD with respect to the local map update parameter $\gamma$.}
    \label{fig:gamma_study}
\end{figure}

In Fig.~\ref{fig:gamma_study}, we study the sensitivity of DRO-GD with respect to the $\gamma$ parameter used in the local map update.
One can see that the method's accuracy is not very sensitive to $\gamma$.
We also analyze the estimated side velocity bias and empirical evidence that, given the assumption made in our formulation, it depends on the observed environment in Appendix~\ref{app:vy_bias}.
Overall, this ablation study demonstrates that the proposed direct radar registration is sound and robust. 

\subsection{Public automotive benchmarks}

We have benchmarked the proposed method against state-of-the-art frameworks over the Boreas and MulRan datasets.
In both cases, the radar uses a sawtooth modulation pattern.
Thus, none of the results presented in this section include the Doppler-based objective function $\objective{d}$.

\subsubsection{Boreas dataset}

The Boreas dataset \cite{burnett2023boreas} was collected by driving a vehicle equipped with a sensor suite composed of a 3D lidar, a 2D spinning radar (sawtooth pattern), a front-facing camera, and an RTK-GNSS/INS solution for ground-truthing.
The dataset consists of 44 sequences collected over one year along more than $350\, \si{\km}$ of roads in Toronto, Canada.
Accordingly, the data corresponds to various weather conditions ranging from clear summer sky to heavy snow and rain. 
A specificity of Boreas is that most sequences follow a unique route in a suburban environment, allowing one to test an algorithm's robustness with respect to changing conditions.
The Boreas dataset also offers a leaderboard for odometry benchmarking that leverages 13 of the sequences above.
Note that the GNSS/INS ground-truth is not publicly available for these sequences.

After testing our algorithm on 6 sequences with ground-truth, we submitted our results to the leaderboard with DRO-G.
Our algorithm significantly outperforms all the other methods.
We have reported the four best radar-based methods in Table~\ref{tab:boreas}.
The core principle of CFEAR~\cite{adolfsson2021CFEAR} and STEAM-RIO++~\cite{burnett2024continuous} is similar as both methods extract a point cloud from the incoming radar data before performing registration to multiple previous scans.
A difference is that CFEAR first performs motion distortion correction based on the previous estimate and uses point-to-distribution residuals, while STEAM-RIO++ leverages a \ac{gp} motion prior, point-to-point constraints and \ac{imu} data.
CFEAR++ \cite{li2024cfearpp} builds atop CFEAR by embedding the use of the \ac{imu} and some semantic information.

Our results confirm the ability of DRO-G to provide state-of-the-art performances without requiring a radar with a triangular modulation pattern and $\objective{d}$ as long as the environment contains sufficient geometric cues.
In Appendix~\ref{app:boreas}, we provide a table that details the error obtained for each of the 13 sequences.
As expected, there seems to be no correlation between the weather conditions and the translation error.
An interesting observation is that the orientation estimates of DRO-G using solely the gyroscope integration outperform the other inertial-aided methods.
This means that the radar constraints in CFEAR++ and STEAM-RIO++ play a negative role in the orientation estimates.

\begin{table}
    \centering
    \caption{Boreas dataset SE(2) odometry leaderboard.}
    \begin{tabularx}{\columnwidth}{lYY}
        \toprule
        \textbf{Method} & \textbf{Trans. err. [\%]} & \textbf{Rot. err. [${}^\circ/100\mathrm{m}$]} 
        \\
        \midrule
        DRO-G (ours) & \textbf{0.26} & \textbf{0.05}
        \\ 
        CFEAR++ \cite{li2024cfearpp} & 0.51 & 0.14
        \\
        CFEAR \cite{adolfsson2023cfear} & 0.61 & 0.21
        \\
        STEAM-RIO++ \cite{burnett2024continuous} & 0.62 & 0.18
        \\
        \bottomrule
    \end{tabularx}
    \label{tab:boreas}
\end{table}

\subsubsection{MulRan dataset}

The MulRan dataset \cite{kim2020mulran} has been recorded with a lidar, 2D spinning radar, and an \ac{imu} mounted atop a car while driving in 4 different locations in Daejeon and Sejong, South Korea.
This dataset originally targeted the task of place recognition, but the availability of a 6-DoF trajectory ground-truth also enables the benchmarking of odometry and \ac{slam} algorithms.
However, unlike in the previous experiments, the ground-truth is not obtained with an RTK-GNSS but via the optimization of a pose graph that leverages wheel odometry, an optical-fibre gyroscope, and a VRS-GPS.
Additionally, the quality of the \ac{imu} is significantly lower than in the Boreas and our automotive datasets.

Table~\ref{tab:mulran} shows the KITTI odometry errors of the proposed method and variations of CFEAR as reported in~\cite{adolfsson2023cfear}.
As introduced in the previous subsection, CFEAR relies on a scan-to-local-keyframes registration. 
C-FEAR-3 and C-FEAR-3-50 refer to~\cite{adolfsson2023cfear} using a different number of past keyframes in the registration process: 4 and 50, respectively.
The table shows the ability of DRO-G to outperform \ac{sota} methods even with a low quality \ac{imu}.
The level of rotational accuracy highlights the soundness of the proposed integration-only orientation and reiterates the observation from the Boreas leaderboard that radar data can have a detrimental impact on the system's rotation estimate compared to pure integration and a simplistic bias estimation strategy.

Surprisingly, both CFEAR and our method display errors significantly larger with MulRan data than with the Boreas dataset.
We believe that the nature of the ground-truth generation process leads to noisier poses, but also that extrinsic calibration and synchronization are suboptimal.
For example, we observed that the difference between consecutive \ac{imu} data timestamps presents a variation of $\pm35\%$ around the nominal period of $10\,\si{\ms}$ and that there is a constant offset with respect to the other sensors.
In our experiments, we corrected the \ac{imu} synchronization by removing $50\si{\ms}$ to each inertial timestamp of all sequences.
We also adjusted the radar heading by $0.172^\circ$ (without the heading correction DRO-G leads to errors of $1.46\%$ and $0.38\si{\degree/100\m}$).
For completeness, without the gyroscope bias estimation, we obtain errors of $6.00\%$ and $1.57\si{\degree/100\m}$ on average.
Due to the lower quality of the dataset, we believe no strong conclusion can be drawn from this experiment other than the demonstration that DRO-G does not require a high-end gyroscope to perform at the level of \ac{sota} methods.

\begin{table}
    \centering
    \caption{Average error per sequence type of the MulRan dataset.}
    \begin{tabularx}{\columnwidth}{lYYY}
        \toprule
        \textbf{Sequences} & \textbf{CFEAR-3} \scriptsize\cite{adolfsson2023cfear} & \textbf{CFEAR-3-50} \scriptsize\cite{adolfsson2023cfear} & \textbf{DRO-G} \scriptsize(ours)
        \\
        \midrule
        KAIST & 1.65 / 0.70 & 1.53 / 0.68 & \textbf{1.44} / \textbf{0.42} 
        \\ 
        DCC & 1.75 / 0.49 & 1.58 / 0.50 & \textbf{1.54} / \textbf{0.41}
        \\
        Riverside & 1.46 / 0.51 & 1.39 / 0.50 & \textbf{1.38} / \textbf{0.31}
        \\
        \midrule
        Average & 1.62 / 0.57 & 1.50 / 0.56 & \textbf{1.45} / \textbf{0.38}
        \\
        \bottomrule
        \multicolumn{4}{c}{\shortstack[]{\rule{0pt}{2.0ex}\scriptsize KITTI odometry metric reported as \textit{XX / YY} with \textit{XX} [\%] \\\scriptsize and \textit{YY} [$\si{\degree}/100\,\si{\m}$] the translation and orientation errors, respectively.}}
    \end{tabularx}
    \label{tab:mulran}
\end{table}

\subsection{Off-road navigation}
To demonstrate the versatility of the proposed method, we collected a series of radar-inertial data sequences with an off-road mobile robot.

\subsubsection{Dataset description}
Our off-road dataset comprises 16 test sequences divided into four locations ranging progressively from highly structured to unstructured environments.
The robot's maximum velocity is approximately $1.2\,\si{\m/\s}$.
The sensor suite comprises a Navtech RAS3 radar and an Ouster lidar with its embedded \ac{imu}.
The robot is driven over the same loop three or four times for each environment.
The easiest one, \emph{Parking}, takes place in a parking lot with many artificial geometric features such as buildings, fences, and parked cars.
The second location \emph{Industrial} takes place around a large industrial building with part of the trajectory going over a grass patch.
This environment contains a combination of geometric structures (fences and buildings) and sparse open areas (trees and bushes) at the back of the industrial building. 
For the third trajectory, \emph{Grass}, the robot moves over a grass field and occasionally can observe a metallic fence.
The last and most challenging environment, \emph{Woods}, is entirely off-road in the woods.
Fig.~\ref{fig:teaser} (right) shows our Clearpath Warthog robot within the \emph{Woods} environment.

\subsubsection{Baseline}
In this experiment, we benchmark DRO-G against a version of CT-R~\cite{are_we_ready_for} with additional inertial constraints as in~\cite{qiao2024radar}.
We denote this ICP-based baseline as CT-RG.
For each CT-RG run, the gyroscope measurements have been corrected with a constant bias estimated from five seconds of data where the robot is static.
CT-RG's parameters have been tuned using held-out \emph{Parking} sequences.

\subsubsection{Results}

\begin{figure*}
    \centering
    \def\subfigdist{0.0cm}
    \def\vdist{0.5cm}
    \def\hdist{0.5cm}
    \def\imgscale{0.90}
    \def\subfigtextsize{\small}
    \begin{tikzpicture}
        \node[inner sep=0, outer sep=0] (parking) {\includegraphics[clip, width=\imgscale\columnwidth, trim=0.2cm 0.2cm 0.2cm 0.2cm]{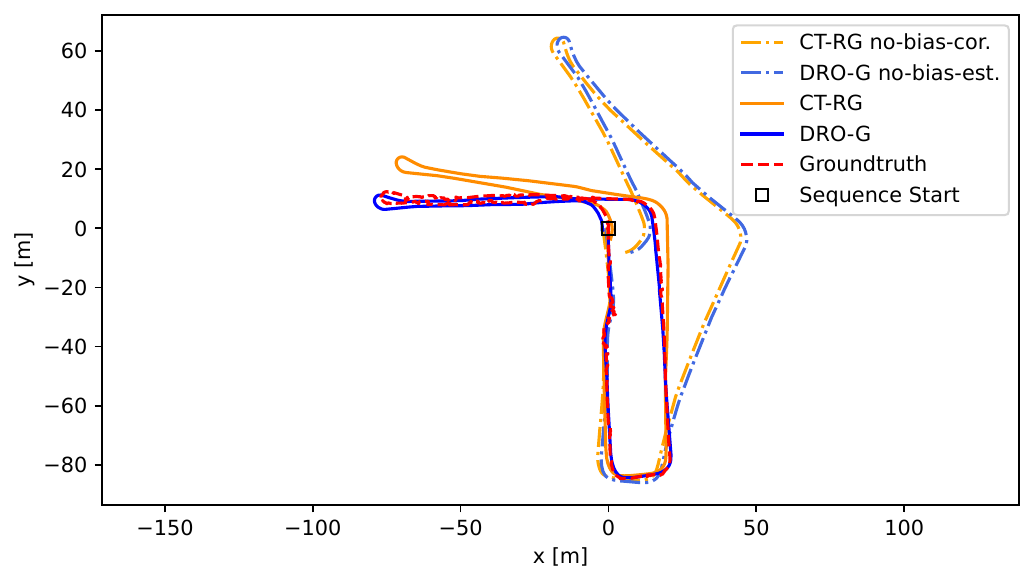}};
        \node[inner sep=0, outer sep=0, right=\hdist of parking] (mars) {\includegraphics[clip, width=\imgscale\columnwidth, trim=0.2cm 0.2cm 0.2cm 0.2cm]{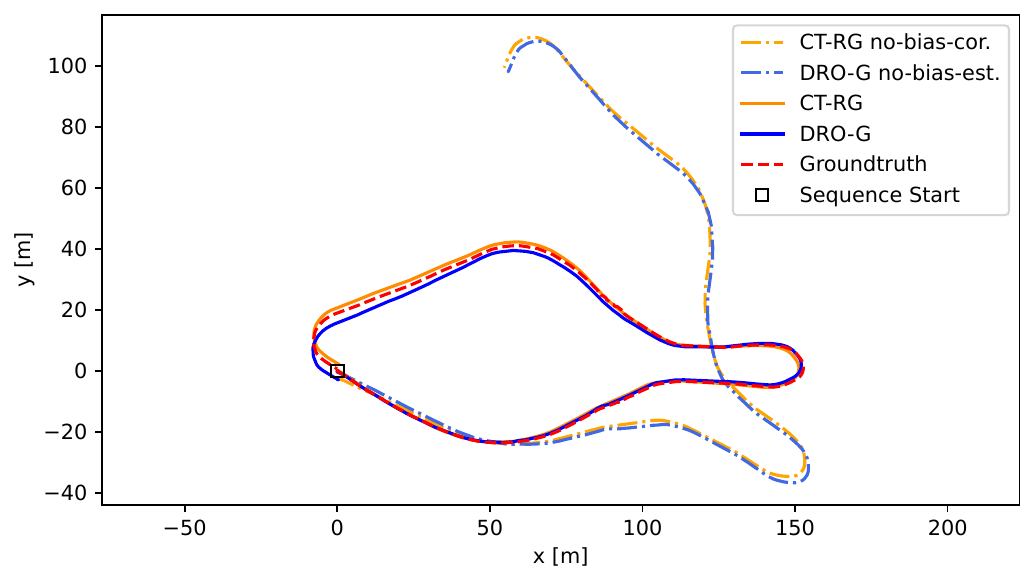}};
        \node[inner sep=0, outer sep=0, below=\vdist of parking] (grass) {\includegraphics[clip, width=\imgscale\columnwidth, trim=0.2cm 0.2cm 0.2cm 0.2cm]{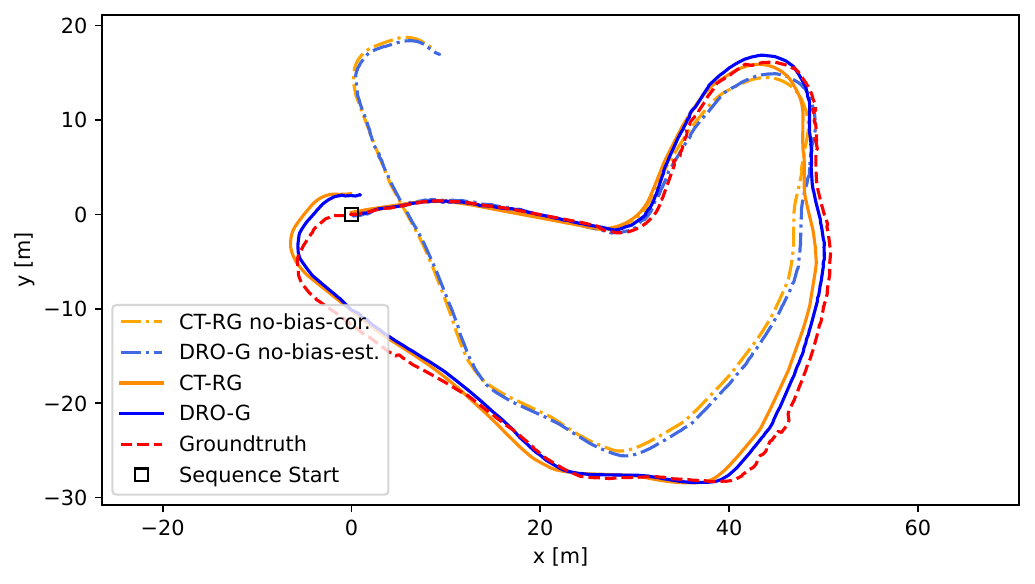}};
        \node[inner sep=0, outer sep=0, right=\hdist of grass] (woods) {\includegraphics[clip, width=\imgscale\columnwidth, trim=0.2cm 0.2cm 0.2cm 0.2cm]{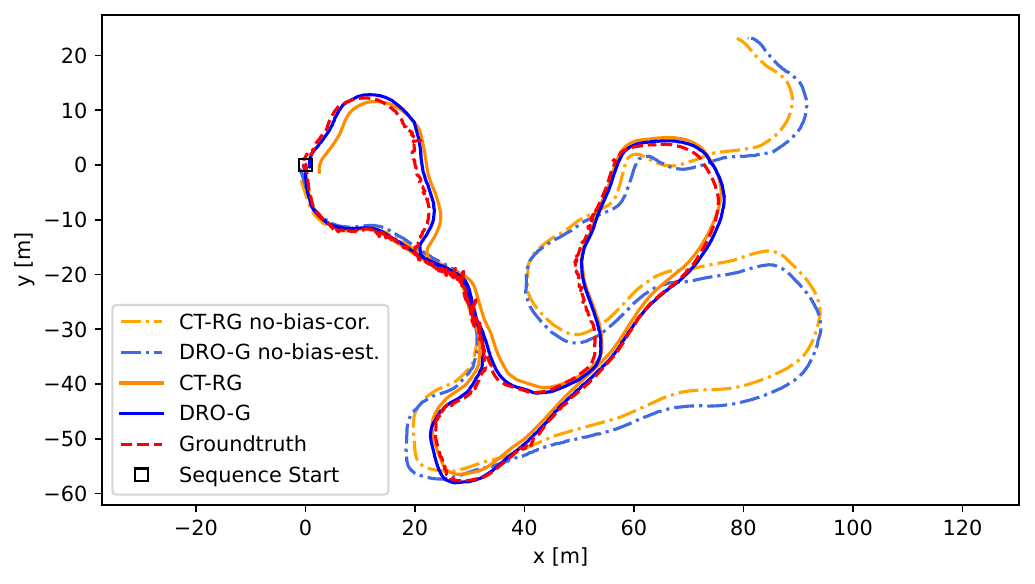}};
        
        \node[inner sep=0, outer sep=0, below=\subfigdist of parking] {\subfigtextsize (a) Parking};
        \node[inner sep=0, outer sep=0, below=\subfigdist of mars] {\subfigtextsize (b) Industrial};
        \node[inner sep=0, outer sep=0, below=\subfigdist of grass] {\subfigtextsize (c) Grass};
        \node[inner sep=0, outer sep=0, below=\subfigdist of woods] {\subfigtextsize (d) Woods};
    \end{tikzpicture}
    \caption{Visualization of trajectory estimates in the off-road scenarios (aligned with the ground-truth for visualization using the first 20\% of the trajectory).}
    \label{fig:offroad_traj}
\end{figure*}
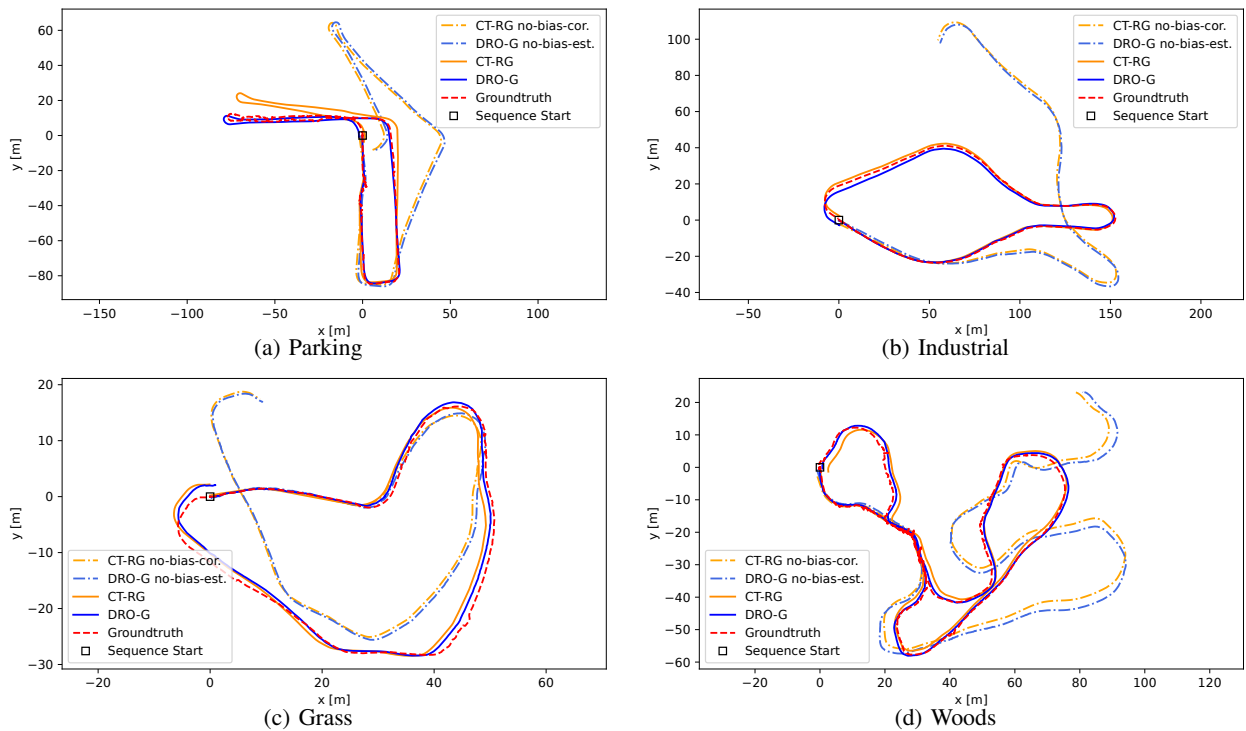

We show the \ac{rpe} obtained with and without gyroscope bias estimation in Table~\ref{tab:warthog}.
In structured environments (\emph{Parking} and \emph{Industrial}), both the direct and point-cloud-based methods perform similarly.
The discrepancy between the two methods increases when dealing with more challenging environments.
Overall, DRO-G maintains a satisfactory level of accuracy across all scenarios, demonstrating that the level of structure of the environment does not seem to impact the accuracy significantly.
It also shows that direct radar registration is robust to the nature of the environment without requiring any tedious parameter tuning.
Fig.~\ref{fig:offroad_traj} provides a visualization of estimated trajectories in each environment.
Table~\ref{tab:warthog} and Fig.~\ref{fig:offroad_traj} also show the efficacy of the proposed gyroscope bias estimation (DRO-G no-bias-est. vs DRO-G).
Unlike the automotive scenario, where the gyroscope bias is nearly negligible, the lower quality of the lidar's \ac{imu} has a significant impact on the results if the bias is not accounted for.
Both methods perform poorly without any bias correction.

\begin{table}
    \centering
    \caption{Average RPE in various environments spanning over different levels of structure.}
    \setlength{\tabcolsep}{2pt}
    \begin{tabularx}{\columnwidth}{lYYYY}
        \toprule
        \textbf{Sequence type} & \textbf{CT-RG}\scriptsize\cite{qiao2024radar} no-bias-est & \textbf{CT-RG}\scriptsize\cite{qiao2024radar} & \textbf{DRO-G} \scriptsize no-bias-est. & \textbf{DRO-G} \scriptsize(ours)
        \\
        \midrule
        Parking \scriptsize($4\times368\,\si{\m}$) & 3.45 & \textbf{1.03} & 3.44 & 1.04
        \\
        Industrial \scriptsize($4\times390\,\si{\m}$) & 3.63 & 0.77 & 3.57 & \textbf{0.62}
        \\ 
        Grass \scriptsize($4\times175\,\si{\m}$)& 5.63 & 1.05 & 5.29 & \textbf{0.67}
        \\
        Woods \scriptsize($4\times421\,\si{\m}$)& 4.70 & 1.30 & 3.42 & \textbf{0.71}
        \\
        \bottomrule
        \multicolumn{5}{c}{\scriptsize Relative position error as a percentage [\%] of the distance travelled.}
    \end{tabularx}
    \label{tab:warthog}
\end{table}

\subsection{Computation time}

All the experiments in this paper were performed using a laptop equipped with an Intel i7-13850HX CPU and an Nvidia RTX 5000 Mobile GPU.
On our automotive dataset, the computation of DRO-GD took on average $89\,\si{\ms}$ per radar frame using the GPU.
This is far below the real-time limit of $250\,\si{\ms}$ ($4\,\si{\Hz}$ radar).
When considering solely the Doppler-based velocity objective function, DRO-D required only $34\,\si{\ms}$ per scan.
For comparison, we also changed the target device to CPU in Pytorch.
We obtained averages of $738\,\si{\ms}$ and $59\,\si{\ms}$ for DRO-GD and DRO-D, respectively.
On the Boreas and MulRan datasets, the GPU computations needed $79\,\si{\ms}$ and $95\,\si{\ms}$ per scan, respectively.
The off-road experiments were the ones that required the most computation time with $134\,\si{\ms}$ per frame.
To summarise, all our experiments have run in real-time.
However, the current implementation is somewhat modular, allowing for different combinations of objective functions and motion models.
We believe that specializing the code and leveraging a compiled programming language would lead to significant improvement in the proposed method's efficiency, especially for CPU-based operations.

\section{Limitations}

Currently, the proposed framework strongly relies on the gyroscope data as the orientation is obtained via direct integration of the angular velocity.
We demonstrated that it was sufficient to obtain \ac{sota} performance in both automotive and off-road scenarios with high-quality and low-quality \acp{imu}.
However, significant degradation of the inertial data (saturation, long dropouts, erroneous measurements, etc), while unlikely, would have a catastrophic impact on the estimated state.
On a similar topic, the proposed approach to estimate the gyroscope bias requires the vehicle/robot to be static occasionally.
Further work is needed to introduce a principled approach to gyroscope bias estimation in DRO.

Our method and experiments focused on spinning \ac{fmcw} radars.
Thus, our findings are limited to this particular type of sensor.
However, we believe that DRO applies to other radar types (e.g. phased arrays) both in 2D and 3D given a sufficient \ac{fov} and access to the intensity returns (unfortunately, most off-the-shelf automotive radars do not provide this information).
A practical limitation of using 3D data is the potential computational load increase if the radar provides much more information than the radars used in this work.

\section{Conclusion}

We have introduced DRO: a direct method for radar-based odometry aided by a gyroscope.
Unlike most of the top-performing methods in the literature, DRO does not extract point clouds or features in the incoming radar scans but directly leverages the intensity information of the data to perform scan-to-local-map registration.
The orientation is obtained by integrating the gyroscope measurements, and the body-centric velocity is estimated via the maximization of one or two objective functions, depending on the radar's frequency modulation.
With a triangular emission pattern, the Doppler-induced shifts in the data make the radial velocities observable, thus enabling odometry in geometrically challenging environments such as featureless tunnels.
Regardless of the radar's signal shape, DRO is the first direct method that accounts in a principled way for both the Doppler-induced and motion distortion of the radar data in the estimation process.
The real-time GPU-based implementation of the proposed method was shown to outperform existing methods on public benchmarks and datasets we have collected both in automotive and off-road environments.
Future works include the integration of a principled way to estimate the gyroscope bias and perform \ac{slam} by estimating the full trajectory and including loop-closure constraints.

\section*{Acknowledgments}

This paper was partially supported by an Ontario Research Fund - Research Excellence grant.
The authors would like to thank Aoran Jiao for his help in collecting the off-road datasets.

\bibliographystyle{plainnat}
\bibliography{references}

\appendices

\section{Results details on Boreas dataset}
\label{app:boreas}

This appendix presents the detailed results obtained on the 13 sequences of the Boreas dataset leaderboard in Table~\ref{tab:boreas_detailed}.

\begin{table}[ht]
    \centering
    \caption{Per-sequence errors of DRO-G on the Boeras dataset leaderboard.}
    \label{tab:boreas_detailed}
    \begin{tabular}{lccc}
        \toprule
        \multicolumn{1}{c}{\textbf{\shortstack{Sequence\\ID}}} & \multicolumn{1}{c}{\textbf{\shortstack{Weather\\conditions}}} & \multicolumn{1}{c}{\textbf{\shortstack{Trans.\\err. [\%]}}} & \multicolumn{1}{c}{\textbf{\shortstack{Rot. err. \\ $[{}^\circ/100\mathrm{m}]$}}}
        \\
        \midrule
        2020-12-04-14-00 & \FilledCloud\Snow & 0.24 & 0.05
        \\ 
        2021-01-26-10-59 & \FilledSnowCloud\Snow\Snow & 0.20 & 0.04
        \\ 
        2021-02-09-12-55 & \FilledSunCloud\Snow & 0.22 & 0.02
        \\ 
        2021-03-09-14-23 & \Sun & 0.23 & 0.06
        \\ 
        2021-04-22-15-00 & \FilledSnowCloud & 0.25 & 0.05
        \\ 
        2021-06-29-18-53 & \FilledRainCloud & 0.36 & 0.02
        \\ 
        2021-06-29-20-43 & \Cloud\HalfSun & 0.44 & 0.04
        \\ 
        2021-09-08-21-00 & \NoSun & 0.30 & 0.02
        \\ 
        2021-09-09-15-28 & \SunCloud & 0.53 & 0.19
        \\ 
        2021-10-05-15-35 & \FilledCloud & 0.12 & 0.02
        \\ 
        2021-10-26-12-35 & \FilledRainCloud & 0.13 & 0.01
        \\ 
        2021-11-06-18-55 & \NoSun & 0.32 & 0.10
        \\ 
        2021-11-28-09-18 & \FilledSnowCloud\Snow\Snow & 0.10 & 0.01
        \\
        \bottomrule
        \multicolumn{4}{c}{\FilledCloud: Overcast, \Snow: Snow coverage, \Snow\Snow: High snow coverage}
        \\
        \multicolumn{4}{c}{ \FilledSnowCloud: Snowing, \Sun: Sun, \FilledRainCloud: Rain, \HalfSun: Dusk, \NoSun: Night}
    \end{tabular}
\end{table}

\section{Lateral velocity bias estimate}
\label{app:vy_bias}

\begin{figure}
    \centering
    \includegraphics[width=0.95\columnwidth]{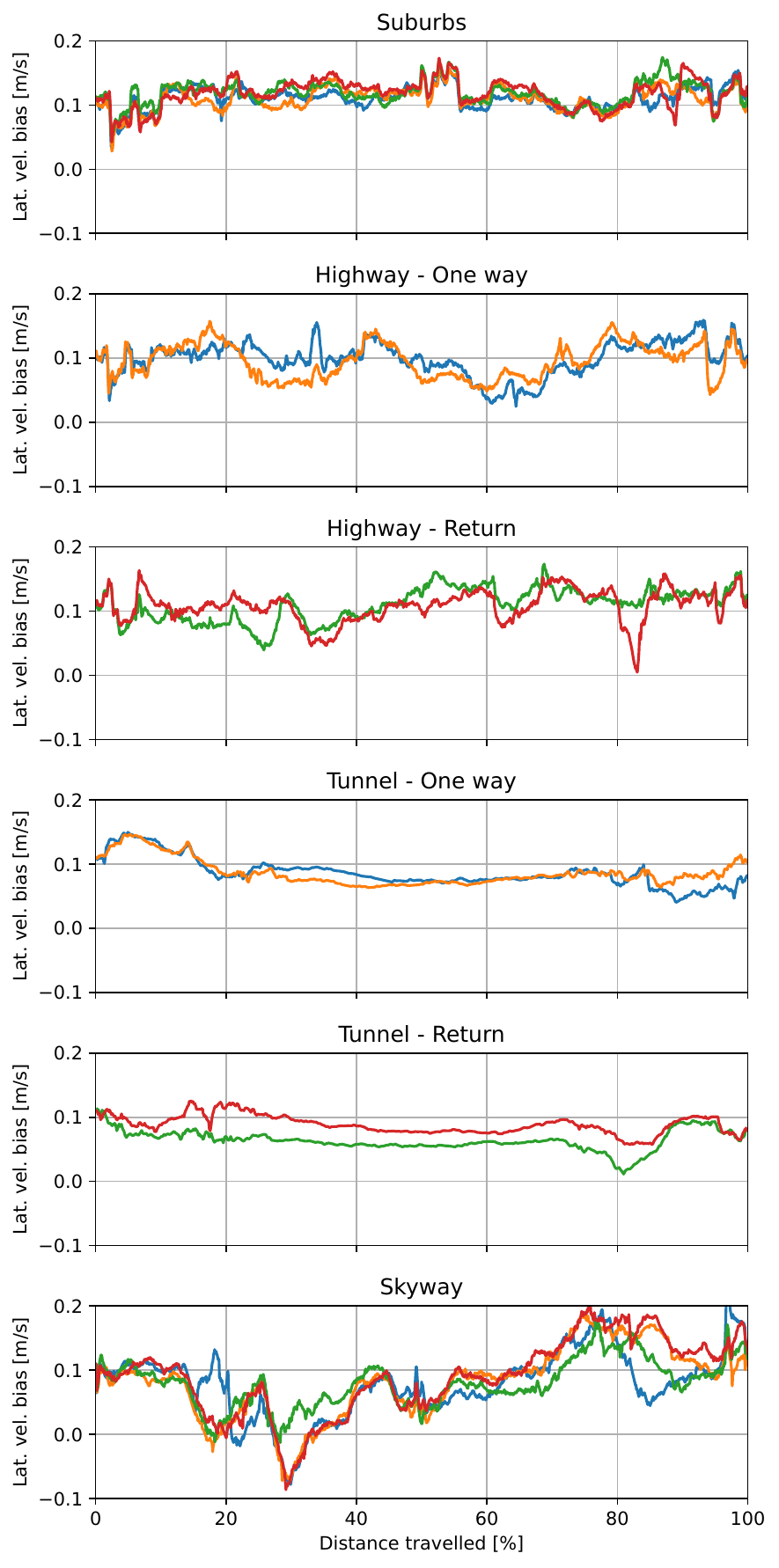}
    \vspace{-0.5cm}
    \caption{Lateral velocity bias estimate obtained by running DRO-GD on our automotive dataset. The bias is expressed as a function of the distance travelled in the sequence.}
    \label{fig:vy_bias}
\end{figure}

This appendix discusses the lateral velocity bias estimation introduced in Section~\ref{sec:vy_bias}.
Fig.~\ref{fig:vy_bias} presents the lateral velocity estimates obtained across the different sequences used in Section~\ref{sec:exp_automotive}.
The value of the bias is shown as a function of the vehicle's position along the trajectory.
Note that the \emph{Highway} and \emph{Tunnel} sequences are split into two plots as they have been collected separately in one direction and the other.
As one can see, there is a clear correlation between the location of the vehicle and the value of the velocity bias estimate.
This observation supports our hypothesis that the Doppler-based velocity bias is impacted by the structure of the environment.
Note that the extrinsic calibration between the radar and the vehicle rear axle has been thoroughly verified, and thus, the miscalibration hypothesis has been discarded.
Further work will be required to obtain a deeper characterization of this bias and a clearer understanding of the physical phenomenon behind it.

\end{document}

%% file: macros.tex
\usepackage[dvipsnames]{xcolor}
\usepackage{moreverb,url}
\usepackage{amsmath,amsfonts,amssymb}
\usepackage{tikz,tkz-euclide}
\usepackage{balance}
\usetikzlibrary{decorations,calligraphy}
\usetikzlibrary{calc}
\tikzset{>=latex}
\usepackage{multicol}
\usepackage{tabularx}
\usepackage{booktabs}
\usepackage{siunitx}
\newcolumntype{Y}{>{\centering\arraybackslash}X}

\usepackage[nolist,nohyperlinks]{acronym}

\def\m{m}

\def\identitymat{\mathbf{I}}

\def\reals{\mathbb{R}}

\def\gp{\mathcal{GP}}
\def\gaussian{\mathcal{N}}

\def\ordinate{y}
\def\ordinatevec{\mathbf{y}}

\def\abscissavec{\mathbf{x}}
\def\abscissamat{\mathbf{X}}

\newcommand{\kernel}[3]{{{k}_{#1}\left(#2,#3\right)}}
\newcommand{\kernelmat}[3]{{\mathbf{K}_{#1}{}_{#2#3}}}

\newcommand{\kernelvec}[3]{{\mathbf{k}_{#1}{}_{#2#3}}}

\newcommand\imutime[1]{{\mathfrak{t}_{#1}}}

\newcommand\objective[1]{{O_{#1}}}
\def\state{{\mathcal{S}}}

\newcommand\rot[2]{{\mathbf{R}_{#1}^{#2}}}

\newcommand\pos[2]{{\mathbf{p}_{#1}^{#2}}}

\newcommand\vel[2]{{\mathbf{v}_{#1}^{#2}}}

\def\world{{W}}

\newcommand\angvel[1]{{\omega_{#1}}}

\newcommand\gyr[1]{{\tilde{\omega}_{#1}}}

\def\noise{{\eta}}

\newcommand\az[1]{{\alpha_{#1}}}
\newcommand\range[1]{{\hat{r}_{#1}}}
\newcommand\intensity[1]{{\psi_{#1}}}
\def\up{\uparrow}
\def\down{\downarrow}
\newcommand\intensityup[1]{{\psi^{\up}_{#1}}}
\newcommand\intensitydown[1]{{\psi^{\down}_{#1}}}
\newcommand\image[1]{{\mathcal{I}^{#1}}}
\newcommand\aztime[1]{{t_{#1}}}
\newcommand\interp[1]{{\Gamma_{#1}}}
\def\linear{\mathcal{L}}
\def\bilinear{\mathcal{B}}
\def\map{\mathcal{M}}

\def\step{c}

\begin{acronym}[AAAAAAAAA]
    \acro{1d}[1D]{One-Dimensional}
    \acro{2d}[2D]{Two-Dimensional}
    \acro{2fast}[2Fast-2Lamaa]{Fast Field-based Agent-Subtracted Tightly-coupled Lidar Localisation And Mapping with Accelerometer and Angular-rate}
    \acro{3d}[3D]{Three-Dimensional}
    \acro{cas}[CAS]{Centre for Autonomous Systems}
    \acro{cnn}[CNN]{Convolutional Neural Network}
    \acro{cpu}[CPU]{Central Processing Unit}
    \acro{doals}[DOALS]{Urban Dynamic Objects LiDAR}
    \acro{dof}[DoF]{Degree-of-Freedom}
    \acro{dvs}[DVS]{Dynamic Vision Sensor}
    \acrodefplural{dvs}[DVS's]{Dynamic Vision Sensors}
    \acro{ekf}[EKF]{Extended Kalman filter}
    \acro{fifo}[FIFO]{First In, First Out}
    \acro{fmcw}[FMCW]{Frequency-Modulated Continuous-Wave}
    \acro{fov}[FoV]{Field-of-View}
    \acro{gnss}[GNSS]{Global Navigation Satellite System}
    \acrodefplural{gnss}[GNSS's]{Global Navigation Satellite Systems}
    \acro{gp}[GP]{Gaussian Process}
    \acrodefplural{gp}[GPs]{Gaussian Processes}
    \acro{gpm}[GPM]{Gaussian Preintegrated Measurement}
    \acro{ugpm}[UGPM]{Unified Gaussian Preintegrated Measurement}
    \acro{gps}[GPS]{Global Position System}
    \acrodefplural{gps}[GPS's]{Global Position Systems}
    \acro{gpu}[GPU]{Graphic Processing Unit}
    \acro{hdr}[HDR]{High Dynamic Range}
    \acro{hri}[HRI]{Human-Robot Interactions}
    \acro{icp}[ICP]{Iterative Closest Point}
    \acro{idol}[IDOL]{IMU-DVS Odometry using Lines}
    \acro{iou}[IoU]{Intersection over Union}
    \acro{imu}[IMU]{Inertial Measurement Unit}
    \acro{in2laama}[IN2LAAMA]{INertial Lidar Localisation Autocalibration And MApping}
    \acro{kf}[KF]{Kalman Filter}
    \acro{kl}[KL]{Kullback–Leibler}
    \acro{lidar}[LiDAR]{Light Detection And Ranging Sensor}
    \acro{lpm}[LPM]{Linear Preintegrated Measurement}
    \acro{map}[MAP]{Maximum A Posteriori}
    \acro{mle}[MLE]{Maximum Likelihood Estimation}
    \acro{mmwave}[mmWave]{millimeter wave}
    \acro{ndt}[NDT]{Normal Distribution Transform}
    \acro{pca}[PCA]{Principal Component Analysis}
    \acro{pm}[PM]{Preintegrated Measurement}
    \acro{rcnn}[R-CNN]{Region-based Convolutional Neural Network}
    \acro{rgb}[RGB]{color}
    \acro{rgbd}[RGBD]{color-depth}
    \acro{rms}[RMS]{Root Mean Squared}
    \acro{rmse}[RMSE]{Root Mean Squared Error}
    \acro{ros}[ROS]{Robot Operating System}
    \acro{rpe}[RPE]{Relative Position Error}
    \acro{rrbt}[RRBT]{Rapidly Exploring Random Belief Trees}
    \acro{sde}[SDE]{Stochastic Differential Equation}
    \acro{SE3}[SE(3)]{Special Euclidean group in three dimensions}
    \acro{slam}[SLAM]{Simultaneous Localisation And Mapping}
    \acro{so3}[$\mathfrak{so}$(3)]{Lie algebra of special orthonormal group in three dimensions}
    \acro{SO3}[SO(3)]{Special Orthonormal rotation group in three dimensions}
    \acro{sota}[SOTA]{state-of-the-art}
    \acro{upm}[UPM]{Upsampled-Preintegrated-Measurement}
    \acro{uts}[UTS]{University of Technology, Sydney}
    \acro{vi}[VI]{Visual-Inertial}
    \acro{vio}[VIO]{Visual-Inertial Odometry}
    \acro{vo}[VO]{Visual Odometry}
\end{acronym}